%% file: main.tex
\documentclass[sigconf]{acmart}
\makeatletter
\def\@ACM@checkaffil{% Only warnings <<<<<<<<<<<<<<<<
    \if@ACM@instpresent\else
    \ClassWarningNoLine{\@classname}{No institution present for an affiliation}%
    \fi
    \if@ACM@citypresent\else
    \ClassWarningNoLine{\@classname}{No city present for an affiliation}%
    \fi
    \if@ACM@countrypresent\else
        \ClassWarningNoLine{\@classname}{No country present for an affiliation}%
    \fi
}
\makeatother

\AtBeginDocument{%
  \providecommand\BibTeX{{%
    \normalfont B\kern-0.5em{\scshape i\kern-0.25em b}\kern-0.8em\TeX}}}

\usepackage{multicol}
\usepackage{multirow}
\usepackage{tablefootnote}
\usepackage{siunitx}
\usepackage{soul}
\usepackage{xcolor}

\usepackage[super]{nth}

\usepackage{tikz}
\usepackage{xcolor}

\newcommand{\ballnumber}[1]{\tikz[baseline=(myanchor.base)] \node[circle,fill=.,inner sep=1pt] (myanchor) {\color{-.}\bfseries\footnotesize #1};}
\newcommand{\blueballnumber}[1]{\tikz[baseline=(myanchor.base)] \node[circle,fill=blue!80,inner sep=1pt] (myanchor) {\color{-.}\bfseries\footnotesize #1};}
\usepackage{tabularx}
\newcolumntype{M}[1]{>{\centering\arraybackslash}m{#1}} % centered
\newcolumntype{R}[1]{>{\raggedleft\arraybackslash}m{#1}} % right centered
\newcolumntype{L}[1]{>{\raggedright\arraybackslash}m{#1}} % right centered
\usepackage{booktabs}

\usepackage{arydshln}

\usepackage{wrapfig}
\usepackage{graphicx}
\usepackage{subcaption}
\usepackage{amsmath}
\usepackage[ruled,vlined]{algorithm2e}

\copyrightyear{2023} 
\acmYear{2023} 
\setcopyright{acmlicensed}\acmConference[SC '23]{The International Conference for High Performance Computing, Networking, Storage and Analysis}{November 12--17, 2023}{Denver, CO, USA}
\acmBooktitle{The International Conference for High Performance Computing, Networking, Storage and Analysis (SC '23), November 12--17, 2023, Denver, CO, USA}
\acmPrice{15.00}
\acmDOI{10.1145/3581784.3607045}
\acmISBN{979-8-4007-0109-2/23/11}

\begin{document}

\title{Co-design Hardware and Algorithm for Vector Search}

\author{Wenqi Jiang$^{1}$, Shigang Li$^{1}$, Yu Zhu$^{1}$, Johannes de Fine Licht$^{1}$, Zhenhao He$^{1}$,  Runbin Shi$^{1}$, Cedric Renggli$^{2}$, Shuai Zhang$^{1}$, Theodoros Rekatsinas$^{2}$, Torsten Hoefler$^{1}$, Gustavo Alonso$^{1}$}
\affiliation{%
  \institution{$^{1}$ETH Zurich\hspace{1em}  $^{2}$Apple}
  % \institution{$^{1}$ETH Zurich\hspace{1em} $^{2}$Beijing University of Posts and Telecommunications\hspace{1em}  $^{3}$Apple}
  % \institution{}
%   \city{Zurich}
%   \country{Switzerland}
}
%   \institution{Systems Group, ETH Zurich}
% \email{{firstname.lastname}@inf.ethz.ch}

\renewcommand{\shortauthors}{Wenqi Jiang et al.}

\begin{abstract}

% Vector search has become the backbone of many ML-powered applications, such as recommender systems and search engines.  
% For example, a search engine can encode a natural language query as a vector and retrieve relevant web pages whose vector representations are similar to the query through vector search. 
Vector search has emerged as the foundation for large-scale information retrieval and machine learning systems, with search engines like Google and Bing processing tens of thousands of queries per second on petabyte-scale document datasets by evaluating vector similarities between encoded query texts and web documents. As performance demands for vector search systems surge, accelerated hardware offers a promising solution in the post-Moore's Law era. We introduce \textit{FANNS}, an end-to-end and scalable vector search framework on FPGAs. Given a user-provided recall requirement on a dataset and a hardware resource budget, \textit{FANNS} automatically co-designs hardware and algorithm, subsequently generating the corresponding accelerator. The framework also supports scale-out by incorporating a hardware TCP/IP stack in the accelerator. \textit{FANNS} attains up to 23.0$\times$ and 37.2$\times$ speedup compared to FPGA and CPU baselines, respectively, and demonstrates superior scalability to GPUs, achieving 5.5$\times$ and 7.6$\times$ speedup in median and 95\textsuperscript{th} percentile (P95) latency within an eight-accelerator configuration. The remarkable performance of \textit{FANNS} lays a robust groundwork for future FPGA integration in data centers and AI supercomputers.

\end{abstract}




\maketitle



% \keywords{datasets, neural networks, gaze detection, text tagging}


% % % \input{new_contents}

\input{introduction}

\input{background}

\input{design_space}

\input{system_overview}

\input{hardware_design}

\input{e2e}

\input{evaluation}

% \input{discussion}

\input{related_work}

\input{conclusion}

\begin{acks} 
We would like to thank AMD for their generous donation of the Heterogeneous Accelerated Compute Clusters (HACC) at ETH Zurich (\url{https://systems.ethz.ch/research/data-processing-on-modern-hardware/hacc.html}), on which the FPGA experiments were conducted. 
% The HACC cluster is publically available to academic researchers: . 
\end{acks}

\bibliographystyle{ACM-Reference-Format}
\balance
\bibliography{ref}

\input{appendix}

% \appendix


\end{document}

%% file: introduction.tex
\section{Introduction}
% \citet{de2020transformations}

% {\color{red} TODO: claim that this is for deployment - recall is fixed across different settings.}

Vector search has emerged as the cornerstone of large-scale information retrieval and machine learning systems. Commercial search engines like Google and Bing process tens of thousands of search queries per second on hundreds of petabytes of data~\cite{Bing_RocksDB, Google_QPS}. These engines employ machine learning models to represent web pages as vectors and retrieve pages by calculating the distances between query vectors and web vectors~\cite{chen2021spann, facebook_EBR, xiong2020approximate, karpukhin2020dense, khattab2020colbert}. Recommender systems~\cite{google_recommendation, suchal2010full} also use vector representations for products to identify and recommend items that users may find appealing.
Recently, large language models (LLMs) have incorporated vector search to enhance model quality. By inputting relevant documents from a vector database~\cite{guu2020realm, lewis2020retrieval, borgeaud2021improving}, LLMs can achieve better performance in various tasks such as open domain question answering and language generation. Vector search is also extensively used in scientific computing, including chemical structure analysis~\cite{bajusz2015tanimoto, willett2014calculation} and biomedical data retrieval~\cite{woodbridge2016improving, neumann2019scispacy, anagnostou2020approximate}.

\begin{figure}[t]
% 	\vspace*{-1em} % to shrink gap between figures
	% \vspace*{-5mm} % to shrink gap between figures
	% full width, can be adjusted
	\centering
  \includegraphics[width=1.0\linewidth]{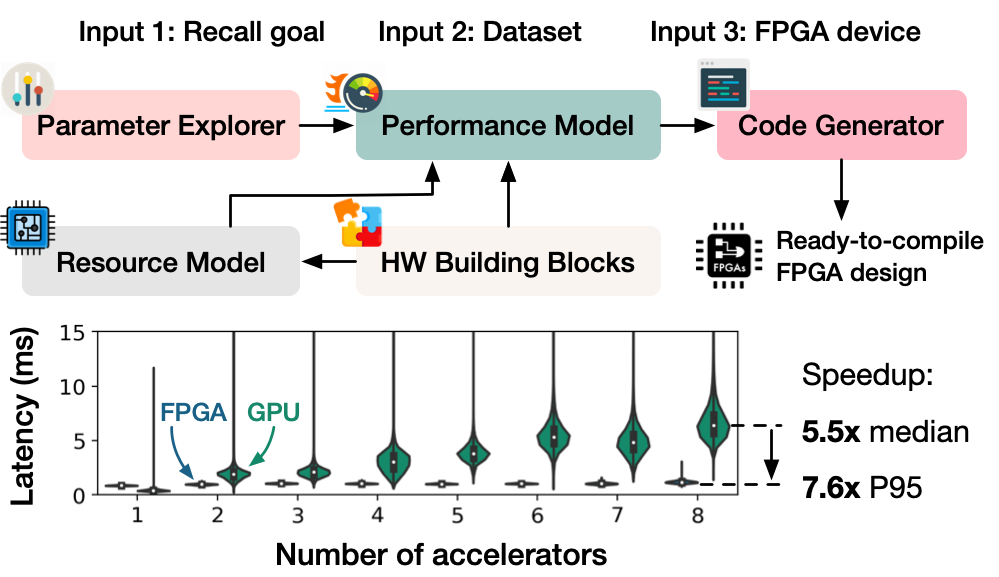}
	\vspace{-1.5em} % to shrink gap between figures
  \caption{{\textit{FANNS} co-designs the hardware and algorithm for vector search. The generated FPGA-based accelerators outperform GPUs significantly in scale-out experiments.}}
  \label{fig:first_page_overview}
	\vspace{-1em} % to shrink gap between figures
\end{figure}

% image retrieval~\cite{SIFT, jegou2011searching}, plagiarism detection~\cite{bandara2011machine}, etc.

% Vector search is the backbone of many ML-powered applications. 
% For example, search engines~\cite{chen2021spann, facebook_EBR, xiong2020approximate} encode web pages as vectors using machine learning models and retrieve pages by evaluating the distances between query vectors and web vectors~\cite{karpukhin2020dense, khattab2020colbert}. 
% Recommender systems~\cite{google_recommendation, suchal2010full} encode products as vectors and retrieve candidate products that users may be interested in by vector search. 
% Vector search is also widely used in image retrieval~\cite{SIFT, jegou2011searching}, language model training~\cite{guu2020realm, borgeaud2021improving}, chemical structure analysis~\cite{wang2021milvus}, etc.

To meet the surging performance demands of vector search systems in the post-Moore's Law era, designing specialized vector search hardware is a promising direction to explore.
% to minimize energy and resource consumption as well as to have flexibility on the deployment. 
Accordingly, we study the hardware specialization of the widely-used IVF-PQ vector search algorithm~\cite{PQ}.
% as it has shown great success on CPUs and GPUs~\cite{matsui2018survey,johnson2019billion}. 
The algorithm uses an inverted file (IVF) index to group vectors into many vector lists by clustering. It then applies product quantization (PQ) to compress high-dimensional vectors into a series of byte codes, reducing memory consumption and accelerating the similarity evaluation process. 
When a query arrives, IVF-PQ goes through six search stages to retrieve similar vectors. The main stages include comparing the vector with all the vector list centroids to identify a subset of relevant lists, scanning the quantized vectors within the selected lists, and collecting the topK most similar vectors. 

% goes through six search stages (e.g., index traversal, vector scanning, and topK collection) to retrieve similar vectors. 
% The highly parallel computation pattern of IVF-PQ also aligns very well with FPGAs. 

\textbf{The benefit of hardware-algorithm co-design.} 
Maximizing the performance of an IVF-PQ accelerator is challenging because one needs to carefully balance the design choices of both the hardware and the algorithm.
Given a target chip size, there are many valid designs to implement IVF-PQ: how should we choose the appropriate microarchitecture for each of the six IVF-PQ search stages? How should we allocate the limited hardware resources to the six stages? 
From the algorithm's perspective, the multiple parameters in IVF-PQ can significantly influence performance bottlenecks and recall.
Due to the vast design space, hardware specialization tailored to a specific set of algorithm parameters can achieve better performance-recall trade-offs: as we will show in the experiments, the accelerators without algorithm parameter awareness are 1.3$\sim$23.0$\times$ slower than the co-designed accelerators given the same recall requirement. 

\textbf{Proposed solution.} 
Considering the numerous design possibilities for an IVF-PQ accelerator, we exploit the reconfigurability of FPGAs to examine various design points. \textbf{We propose \textit{FANNS} (FPGA-accelerated Approximate Nearest Neighbor Search), an end-to-end accelerator generation framework for IVF-PQ-based vector search, which automatically co-designs algorithm and hardware to maximize the accelerator performance for target datasets and deployment recall requirements.} 
Figure~\ref{fig:first_page_overview} illustrates the FANNS workflow. 
Provided with a dataset, a deployment recall requirement, and a target FPGA device, \textit{FANNS} automatically (a) identifies the optimal combination of parameter settings and hardware design and (b) generates a ready-to-deploy accelerator. 
Specifically, \textit{FANNS} first evaluates the relationship between IVF-PQ parameters and recall on the given dataset. 
It also lists all valid accelerator designs given the FPGA hardware resource budget.
Then, the \textit{FANNS} performance model predicts the queries-per-second (QPS) throughput of all combinations between algorithm parameters and hardware designs.
Finally, using the best combination determined by the performance model, the \textit{FANNS} code generator creates the corresponding FPGA code, which is compiled into an FPGA bitstream. 
Besides a single-accelerator solution, \textit{FANNS} can also scale out by instantiating a hardware TCP/IP stack in the accelerator.

\textbf{Results.} 
% We evaluate \textit{FANNS} on the SIFT dataset given various recall requirements. 
Experiments conducted on various datasets demonstrate the effectiveness of the hardware-algorithm co-design: the accelerators generated by \textit{FANNS} achieve up to 23.0$\times$ speedup over fixed FPGA designs and up to 37.2$\times$ speedup compared to a Xeon CPU. While a GPU may outperform an FPGA due to its higher flop/s and bandwidth, FPGAs exhibit superior scalability compared to GPUs thanks to the stable hardware processing pipeline. As shown in Figure~\ref{fig:first_page_overview}, experiments on eight accelerators show that the FPGAs achieve 5.5$\times$ and 7.6$\times$ speedup over GPUs in median and 95\textsuperscript{th} percentile (P95) latency, respectively. 

% The experiments also indicate that future ASIC designs should avoid the problem of shifting bottlenecks by targeting only a subset of search operators and cooperating with a general-purpose processor. 
% {\color{red} Moreover, we extend the FPGAs with a TCP/IP network stack to support scale out. Our solution running on eight FPGAs show xxx better scalability than the state-of-the-art GPU solution. }

% ; and the customized accelerator designs look very different given various recall requirements.
% Normalizing QPS by computation capacity (flop/s), \textit{FANNS} shows up to 3.22$\times$ speedup over GPUs, demonstrating its advantages in settings where a GPU is not an option from an architectural point of view (e.g., embedded systems, computation at the edge, streaming architectures, smart-NICs, near-storage or near-memory processing, etc.). 

% The performance model used to develop FANNS is proven effective as the evaluated throughput reaches 95.3\% of the model's prediction.

% \textit{To the best of our knowledge, this is the first paper that comprehensively studies how to design specialized hardware for ANNS by exploring the huge design space.} 
\textbf{Contributions}
	\vspace{-1em}
\begin{itemize}%[wide = 0pt]
        \item We identify a major challenge in designing accelerators for the IVF-PQ-based vector search algorithm: handling the shifting performance bottlenecks when applying different algorithm parameters.
	\item We show the benefit of co-designing hardware and algorithm for optimizing large-scale vector search performance.
	% \vspace{-0.5em}
	% \item We propose \textit{FANNS}, a hardware-algorithm co-design framework for vector search, which includes: 
        \item We propose \textit{FANNS}, an end-to-end accelerator generation framework for IVF-PQ, maximizing accelerator performance for target datasets and recall requirements. FANNS includes:
	% \vspace{-0.5em}
	\begin{itemize}%[wide = 0pt]
	    % \vspace{-0.5em}
		\item A collection of hardware building blocks for IVF-PQ.
		\item An index explorer that captures the relationship between algorithm parameters and recall. 
		\item A hardware resource consumption model that returns all accelerator designs on a given FPGA device.
		\item A performance model to predict the accelerator QPS of arbitrary combinations of algorithm parameters and accelerator designs.
		\item A code generator that creates ready-to-compile FPGA code given arbitrary accelerator designs. 
		% \item An TCP/IP network stack to support system scale-out. 
	\end{itemize}
	% \vspace{-0.5em}
	\item{We demonstrate the impressive performance and scalability of \textit{FANNS}, achieving 7.6$\times$ P95 latency speedup over GPUs when utilizing eight accelerators.}
	
\end{itemize}

%% file: background.tex
\section{Background}
\label{sec:background}

% \subsection{Vector Search Problem Definition} 

A vector search takes a query vector $x\in \mathcal{R}^{d}$ as input and retrieves $K$ relevant vector(s) (based on, e.g., L2 distances) from the database $Y$ which contains many $d$-dimensional vectors. While nearest neighbor search retrieves the exact $K$ closest vectors, really world vector search systems adopt approximate nearest neighbor (ANN) search that trades accuracy for much higher search performance (latency and throughput). The quality of ANN is measured by the recall at $K$ ($R@K$). In this paper, we use the terms \textit{vector search} and \textit{ANN search} interchangeably.

%  $Y=\{y_i\}_{i=1,...,n}$ where $y_i\in \mathcal{R}^{d}$

% or NNS, is defined as… Practically, people also use ANN. recall R@K

% Given a $D$-dimensional query vector $q\in {\R}^D$ and a database $\mathcal{Y}=\{ y_1,...,y_\mathcal{N}\} \subset {\R}^D$ containing $\mathcal{N}$ vectors, the goal of a $K$-nearest neighbor search (K-NNS) is to return the $K$ ($K \ll \mathcal{N}$) vectors in a set $R_{K} \subset \mathcal{Y}$ (with $\vert R_{K} \vert =K$) that are the most similar to $q$, based on a distance function $d(q,y)$ (such as Euclidean distance) for all $y \in \mathcal{Y}$. 
% In this paper, we adopt the typically used Euclidean distance to measure the similarity between vectors. 
% Formally, the following property holds for a given query vector $q$:
% \begin{equation}
% \forall r \in R_{K}, \forall y \in \mathcal{Y} \setminus R_{K}:\quad d(q, r) \leq d(q, y).
% \end{equation}

% Various approximate nearest neighbor search (ANNS) algorithms have been proposed to speed up the retrieval at the cost of precision. 
% An ANNS returns a set $\Tilde{R}_{K}$ which is approximately equal to the \emph{true} NNS result $R_{K}$. We measure the result quality of the ANNS by reporting the recall $R@K$, which represents the percentage of the $K$ elements in $\Tilde{R}_{K}$ that are also contained in $R_{K}$.

% ANNS analytical database systems

\subsection{The IVF-PQ Algorithm}
\label{sec:ivfpq}

We target IVF-PQ because it is one of the most popular algorithms for large-scale ANN~\cite{PQ}. Its key elements include (a) partitioning the vectors by the inverted-file index to reduce the percentage of vectors to scan and (b) applying product quantizing to compress the vectors and save memory bandwidth. %and (b) approximating rather than precisely computing distances between query vectors and database vectors to reduce computation. 

% \begin{figure}[t!]
% 	% \vspace*{-5mm} % to shrink gap between figures
% 	% full width, can be adjusted
% 	\centering
%   \includegraphics[width=0.95\linewidth]{fig/ANNS-all.png}
%   \caption{Training and querying steps of IVF-PQ.}
%   \label{fig:ANNS-all}
% 	% \vspace*{-5mm} % to shrink gap between figures
% \end{figure}

\begin{table}[t]
\begin{small}
  \begin{center}
    \caption{Definitions of some important symbols.}
    \vspace*{-1em}
    \label{tab:table1}
    \begin{tabular}{L{4.5em} L{19em}} % <-- Alignments: 1st column left, 2nd middle and 3rd right, with vertical lines in between
      \toprule
      Symbol & \multicolumn{1}{c}{Definition} \\
      \midrule
      $x$ & A query vector.\\
      $y$ & A database vector. \\
      $K$ & The number of most similar vectors to return. \\
      $m$ & The sub-space number of product quantization.  \\
      $nlist$ & The totol Voronoi cell number. \\
      $nprobe$ & The number of cells to be scanned per query. \\
      \bottomrule
    \end{tabular}
  \end{center}
 \end{small}
	% \vspace*{-2em} % to shrink gap between figures
\end{table}

\subsubsection{Inverted File (IVF) Index} 
% \hfill \\

An IVF index partitions a vector dataset $Y$ to many ($nlist$) disjoint subsets by clustering algorithms such as k-means. Each partition is known as a Voronoi cell. At query time, only a few ($nprobe$) Voronoi cells close to the query vector are scanned, such that the scanning workloads are reduced. 

% Both the number of Voronoi cells ($nlist$) and the number of scanned cells ($nprobe$) can be (empirically or experimentally) configured to meet various performance and recall requirements.

% As shown in Figure~\ref{fig:ANNS-all}~\ballnumber{1}, the IVF index reduces the number of vectors to be scanned during the search process. At the training stage, a k-means clustering is performed on the input vectors, such that each vector is associated with its closest centroid $C(y)\in {R}^D$. Each cluster is known as a Voronoi cell, and all these cells compose the IVF index. At query time, only a subset of the Voronoi cells are scanned. Both the number of Voronoi cells ($nlist$) and the number of scanned cells ($nprobe$) can be (empirically or experimentally) configured to meet various performance and recall requirements. The IVF index is also known as the coarse-grained quantizer which maps each vector to a Voronoi cell ID.

\subsubsection{Product Quantization (PQ)}
% \hfill \\

\begin{figure}[t]
% 	\vspace*{-1em} % to shrink gap between figures
	% \vspace*{-5mm} % to shrink gap between figures
	% full width, can be adjusted
	\centering
  \includegraphics[width=1.0\linewidth]{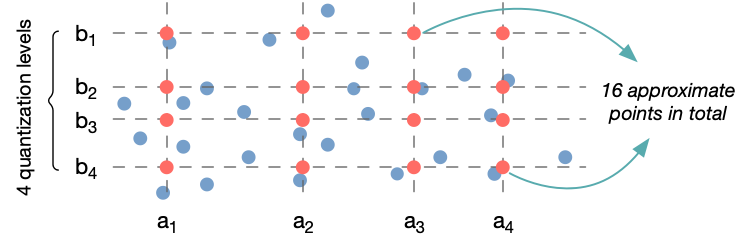}
	\vspace*{-2em} % to shrink gap between figures
  \caption{Product quantization on two-dimensional vectors.}
  \label{fig:PQ}
	\vspace*{-1em} % to shrink gap between figures
\end{figure}

Product quantization (PQ) divides vectors into multiple subvectors and applies quantization per sub-vector space. Figure~\ref{fig:PQ} shows a PQ example on two-dimensional vectors. The example applies four-level quantization per dimension, thus a vector can be approximated by a combination of quantization levels of the two dimensions (e.g., ($a_1$, $b_2$)).  

Practically, PQ can compress a $d$-dimensional vector to $m$-byte ($m<d$) PQ codes. 
Each vector in the dataset is divided into $m$ sub-vectors. 
Within a sub-vector space, all the sub-vectors are clustered into 256 groups, such that a sub-vector can be approximated by its nearest cluster centroid.  
As one can represent the cluster ID (0$\sim$255) by one byte, the entire vector can be stored as $m$-byte PQ codes representing the closest cluster centroids per sub-vector space.

Optionally, one can use optimized product quantization (OPQ) to further improve quantization quality~\cite{OPQ}. The key idea is to rotate the vector space such that the sub-spaces are independent and the variances of each sub-space are normalized. % As a result, OPQ improves the recall over PQ given the same index and the same number of cells to scan.
At query time, OPQ simply introduces a vector-matrix multiplication between the query and the transformation matrix, while the rest search procedures are identical to PQ. 
% \begin{equation}
% \label{eq:opq}
%   E=\sum_{y\in{{Y}}}{||y-Rq(y)|| }
% \end{equation}

% As in Equation~\ref{eq:opq}, OPQ~\cite{OPQ} targets to minimize the distortion of the PQ compression by rotating the vector space by $R$, such that (a) the sub-spaces are independent and (b) the variances of each sub-space are normalized. As a result, OPQ improves the recall over PQ given the same index and the same number of cells to scan.
% \begin{equation}
% \label{eq:opq}
%   E=\sum_{y\in{\mathcal{Y}}}{||y-Rq(y)|| }
% \end{equation}
% From the system's perspective, OPQ is simply a pre-processing operator: at query time, the first step is to perform a vector-matrix multiplication between the query vector $x$ and the OPQ matrix $R$. The rest of the searching process is the same as PQ. 

\subsubsection{The Six Search Stages at Query Time} 
% \hfill \\
IVF-PQ contains six search stages for query serving. 
\textit{First}, if OPQ is involved, transform the query vector by the OPQ matrix (\ul{Stage OPQ}).
\textit{Second}, evaluate the distances between a query vector and all Voronoi cell centroids (\ul{Stage IVFDist}). 
\textit{Third}, select a subset of cells that are closest to the query vector to scan (\ul{Stage SelCell}). 
\textit{Fourth}, in order to compare distances between PQ codes and a query vector efficiently, construct a distance lookup table per Voronoi cell (\ul{Stage BuildLUT}). More specifically, this step divides the query vector into $m$ sub-vectors and computes the distances between the normalized query vector and all centroids of the sub-quantizer.
\textit{Fifth}, approximate the distances between a query vector and the PQ codes (\ul{Stage PQDist}) by Equation~\ref{eq:adc}, in which $d^{2}(x_i,\hat{y_i}$ only requires looking up the distance tables constructed in Stage BuildLUT. This lookup-based distance computation process is also known as asymmetric distance computation (ADC). 
\textit{Finally}, collect the $K$ vectors closest to the query (\ul{Stage SelK}). 

\vspace{-1em}
\begin{equation}
\label{eq:adc}
  \hat{d}^{2}(x,y)=d^{2}(x,\hat{y})
%   =d^{2}(x,[u1(y_1),...um(y_m)])
  =\sum_{i=1}^{m}d^{2}(x_i,\hat{y_i})
\end{equation}
\vspace{-1em}

%% file: design_space.tex
\section{Hardware-Algorithm Design Space}
\label{sec:design_space}

The main challenge in designing compelling IVF-PQ accelerators is to find the optimal option in a huge algorithm-hardware design space, as summarized in Table~\ref{tab:design_space}. 
From the algorithm's perspective, multiple parameters in IVF-PQ can significantly influence recall and performance bottlenecks. 
From the hardware's perspective, there are many valid designs to implement IVF-PQ.
%  summarizes the choices that forms the design space.

\begin{table}%[h]
\begin{small}
  \begin{center}
    \caption{The list of choices during design space exploration.}
    \label{tab:design_space}
    \vspace{-1em}
    \begin{tabular}{L{4.5em} L{19em}} % <-- Alignments: 1st column left, 2nd middle and 3rd right, with vertical lines in between
      \toprule
      \multicolumn{2}{c}{Algorithm parameter space} \\
      \midrule
      \textit{nlist} & The totol Voronoi cell number. \\
      \textit{nprobe} & The number of cells to be scanned per query. \\
      \textit{K} & The number of most similar vectors to return. \\
      \textit{OPQ\textsubscript{enable}} & Whether to apply OPQ. \\
      \midrule
      \multicolumn{2}{c}{Hardware design space} \\
      \midrule
      \textit{Design\textsubscript{s}} & The microarchitecture design of stage $s$. \\
      \textit{\#PE\textsubscript{s}} & The number of processing elements in stage $s$. \\
      \textit{Cache\textsubscript{s}} & Cache index on-chip or store it off-chip for stage $s\in$ \{Stage IVFDist, Stage BuildLUT\}. \\
    %   \textit{Design\textsubscript{s}} & The microarchitecture design of stage $s\in$ \textit{Stages}. \\
    %   \textit{\#PE\textsubscript{s}} & The number of processing elements in stage $s\in$ \textit{Stages}. \\
    %   \textit{Cache\textsubscript{s}} & Cache index on-chip (SRAM) or store index off-chip (DRAM) for stage $s\in$ \{Stage IVFDist, Stage BuildLUT\}. \\
    %   \textit{NoC} & The network-on-chip topology between PEs within a single stage and between two consecutive stages. \\
      \bottomrule
    \end{tabular}
  \end{center}
   \vspace*{-1em} % to shrink gap between figures
\end{small}
\end{table}

\begin{figure*}[t]
  % \vspace*{-5mm} % to shrink gap between figures
  \centering
  
  \begin{subfigure}[b]{0.33\linewidth}
    \includegraphics[width=\linewidth, height=8em]{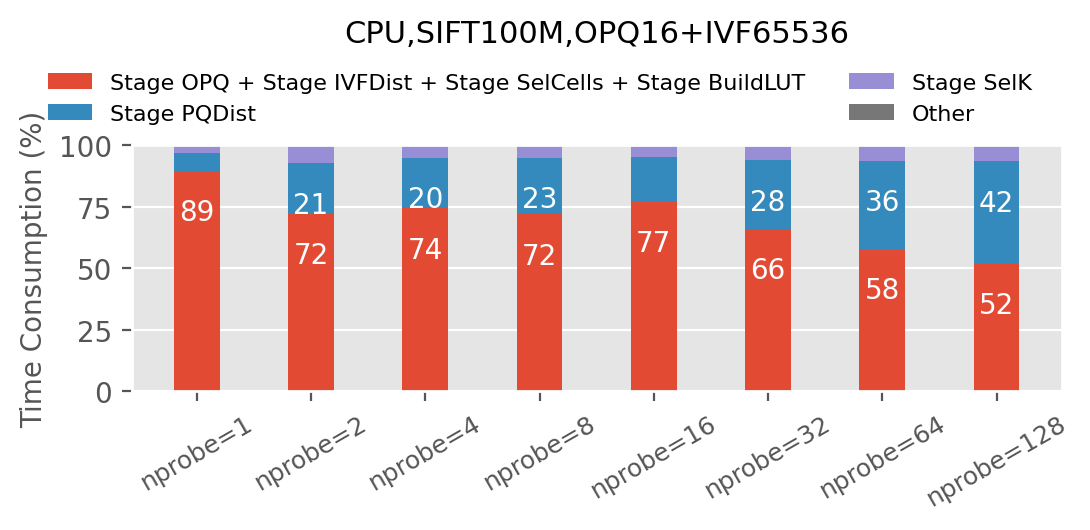}
     % \caption{Coffee.}
  \end{subfigure}
  \hfill
  \begin{subfigure}[b]{0.33\linewidth}
    \includegraphics[width=\linewidth, height=8em]{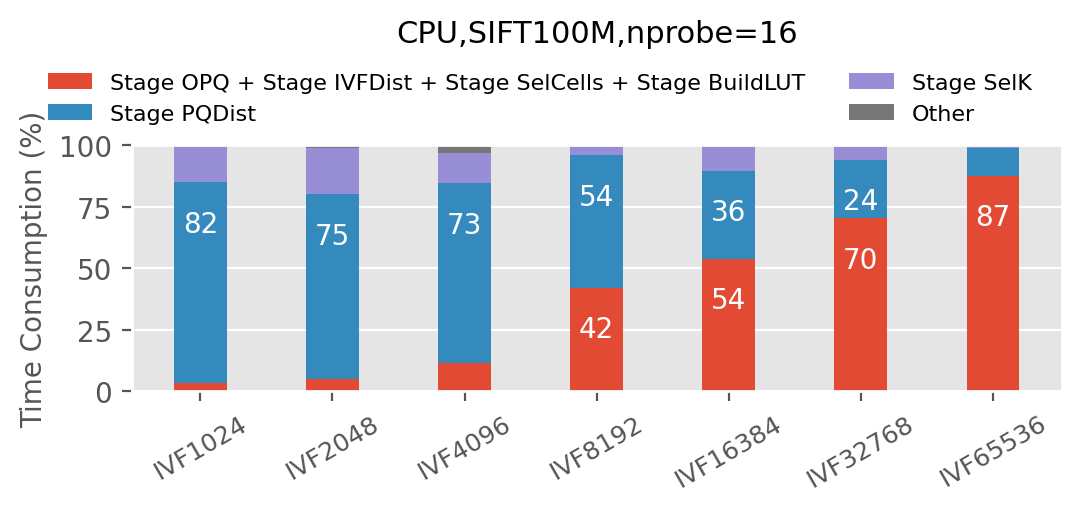}
     % \caption{Coffee.}
  \end{subfigure}
  \hfill
  \begin{subfigure}[b]{0.33\linewidth}
    \includegraphics[width=\linewidth, height=8em]{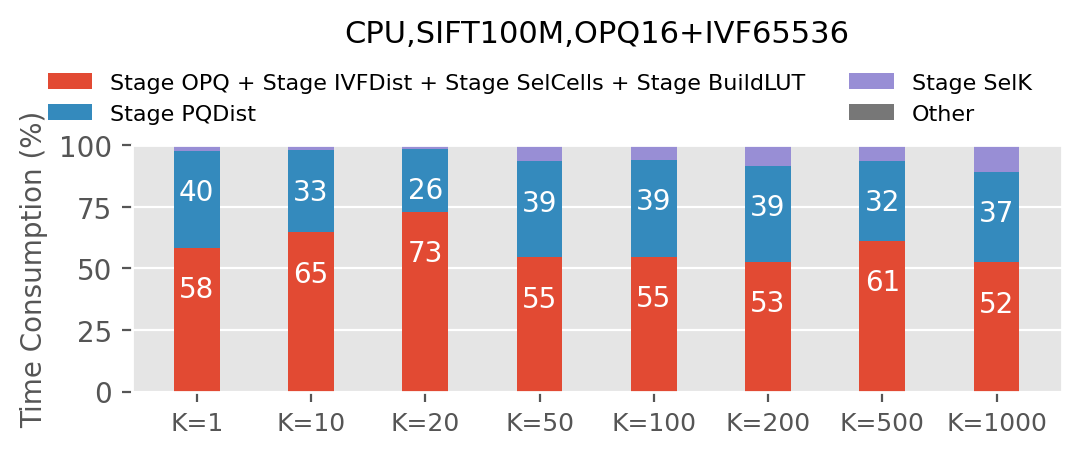}
     % \caption{Coffee.}
  \end{subfigure}

  \begin{subfigure}[b]{0.33\linewidth}
    \includegraphics[width=\linewidth, height=8em]{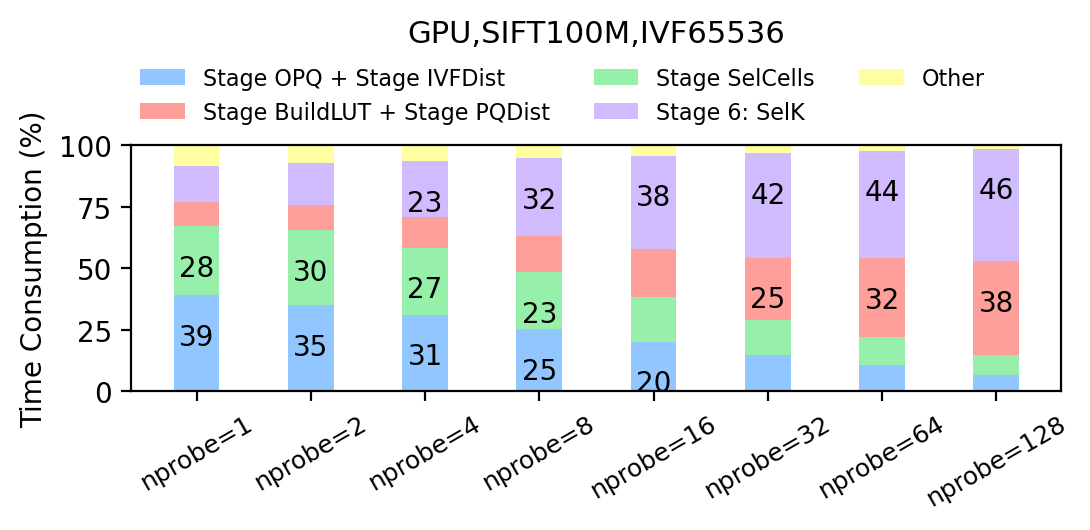}
    % \caption{More coffee.}
  \end{subfigure} 
  \hfill
  \begin{subfigure}[b]{0.33\linewidth}
    \includegraphics[width=\linewidth, height=8em]{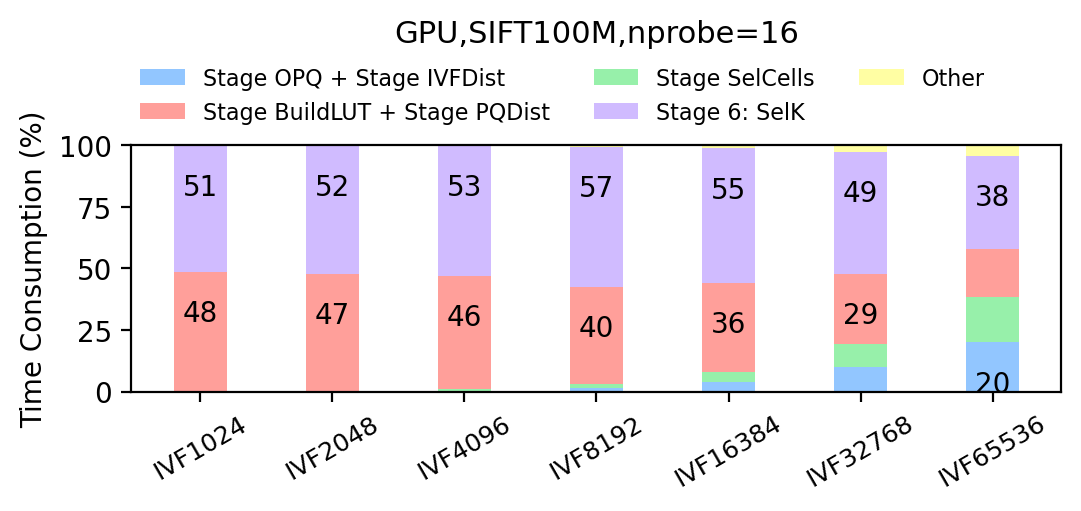}
    % \caption{More coffee.}
  \end{subfigure}
  \hfill
  \begin{subfigure}[b]{0.33\linewidth}
    \includegraphics[width=\linewidth, height=8em]{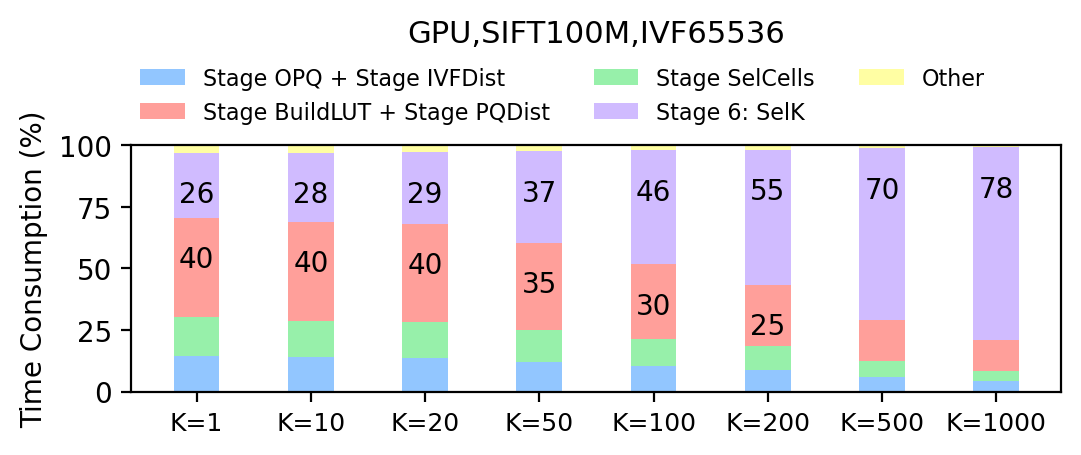}
    % \caption{More coffee.}
  \end{subfigure}
  
  \vspace*{-1em}
  \caption{IVF-PQ bottleneck analysis on CPU (1st row) and GPU (2nd row). By tuning \textit{nprobe} (1st column), \textit{nlist} (2nd column), and $K$ (3rd column), we find that the bottlenecks shift across different algorithm parameters.}
  \label{fig:cpu_gpu_profiling}
  \vspace{-1em}

\end{figure*}

\subsection{Algorithm Parameter Space}
% \hfill \\

\textbf{To achieve a certain recall requirement, there are many options for selecting algorithm parameters.} 
For example, as we will present in the experiments, all the indexes we evaluated can achieve a target recall of R@100=95\% by different \textit{nprobe}. 
It is hard to tell which set of parameters we should deploy on the accelerator.

% \begin{figure}%[h]
%   % \vspace*{-5mm} % to shrink gap between figures
%   % full width, can be adjusted
%   \centering
%   \includegraphics[width=1.0\linewidth]{fig/recall_curve.png}
%   \vspace*{-2em} 
%   \caption{Recall curves of several algorithm settings returning the K=10 most similar vectors.}
%   \vspace{-1em}
%   \label{fig:recall-curve}
% \end{figure}

\textbf{Parameter selections can change the performance bottleneck drastically, which must be considered during the accelerator design phase.} 
We profile the search process on CPUs and GPUs and break down the time consumption per search stage in Figure~\ref{fig:cpu_gpu_profiling}.
% ~\footnote{We count time per stage by adding up the time consumption of all function calls within the stage. We merge search stages when a function is called in multiple stages.}. 
% By tuning one parameter at a time, .
Unlike many applications with a single outstanding bottleneck, the bottlenecks of IVF-PQ shift between the six search stages when different parameters are used.  
% The experiments show that the bottleneck shifts dramatically as the parameters change.
However, a specialized accelerator cannot handle shifting bottlenecks because it contains a certain number of dedicated processing elements (PE) for each stage. 
Thus, the accelerator should either target to achieve acceptable performance running arbitrary algorithm parameters or to achieve optimal performance on a certain parameter setting. 
We now break down the IVF-PQ bottlenecks:

\textit{The performance effect of \textit{nprobe}.} 
We fix the index and tune \textit{nprobe}.
We use the indexes that can achieve the highest QPS of R@100=95\% on the SIFT100M dataset on CPU and GPU, respectively.
As shown in the first column of Figure~\ref{fig:cpu_gpu_profiling}, increasing the number of cells to scan results in more time consumption in Stage PQDist and Stage SelK, regardless of hardware platforms. 
The time consumption of these two stages, on GPUs for example, can increase from 20\% to 80\% as \textit{nprobe} grows.

\textit{The performance effect of \textit{nlist}.} 
By contrast to the first experiment, we now observe the effect of the total number of clusters of the index by fixing the number of clusters to scan (\textit{nprobe=16}). As shown in the second column of Figure~\ref{fig:cpu_gpu_profiling}, higher \textit{nlist} results in more time consumption on Stage IVFDist to evaluate distances between the query vector and cluster centroids. The consumption is more significant on CPUs due to their limited flop/s compared with GPUs, while the main bottlenecks of GPUs are still in later stages even if \textit{nlist} is reasonably large.

\textit{The performance effect of $K$.}
We fix the index per hardware as in the $\mathrm{nprobe}$ experiment. As shown in the third column of Figure~\ref{fig:cpu_gpu_profiling}, the time consumption on Stage SelK on GPUs increases significantly as $K$ grows, while the phenomenon is unobvious on CPUs as the bottlenecks are in other stages.

\subsection{Hardware Design Space}

\textbf{There are many ways to implement an IVF-PQ accelerator, and the design choices are summarized in Table~\ref{tab:design_space}.} 

% {\color{red}Wenqi: in the background section, make sure to introduce what is PE.}

{The \textit{first} choice is the microarchitecture per search stage}. Not only does the processing element (PE) design differ between stages, there are multiple valid designs per stage. For example, Stage SelK collects $K$ nearest neighbors from a series of distance values, which can either be implemented by a hierarchical priority queue consisting of systolic compare-swap units or by a hybrid design involving sorting network and priority queues, as we will show in Section~\ref{sec:hardware_design_space}.

{The \textit{second} choice is chip area allocation across the six search stages, i.e., choosing PE numbers per stage}.  
Due to the finite transistors within a chip, this is a zero-sum game: increasing the number of PEs in one stage implies reducing them in another. 
% To maximize the overall accelerator performance, we should picking appropriate PE numbers such that the performance per stage is balanced, since the QPS is constrained by the slowest stage.

{The \textit{third} decision is about index caching}. Though storing them in off-chip DRAM is the only option for larger IVF indexes, we can decide whether to cache smaller indexes in on-chip SRAM. Caching index guarantees low accessing latency and high bandwidth but increases hardware resource consumptions.

% \ul{The fourth decision is the network-on-chip (NoC) topology.} 
%  The PEs within a stage and between stages should be connected by FIFOs for streaming communications. For the PEs sharing the same input data, the NoC topology can either be the broadcasting scheme or the forwarding scheme.

\subsection{How Does One Choice Influence Others?}

The choices of algorithm parameters will influence the optimal hardware design and vice versa.
Since the relationship between the design choices is intricate, we only convey the intuition here with a couple of examples, while the quantitative model will be presented in later sections. 
First, tuning a single parameter can affect the optimal accelerator design.
Increasing $nlist$ results in more workload in comparing the distances between query vectors and IVF centroids. As a result, more PEs in Stage IVFDist should be instantiated to handle the increasing workload, while fewer PEs can be instantiated in other stages due to the limited chip size. Besides, if the $nlist$ is large enough, caching the IVF index on-chip is not a choice at all, while caching small indexes can be beneficial at the cost of consuming on-chip memory that other PEs could have taken. 
Second, a specific accelerator design has its favorable parameter settings.
Assume the accelerator has a lot of Stage IVFDist PEs, while other stages are naturally allocated with fewer resources.
Such design naturally favors a parameter setting of high $nlist$ and low $nprobe$: the reverse case (low $nlist$ and high $nprobe$) will underutilize the Stage IVFDist PEs yet overwhelming the limited Stage PQDist PEs, resulting in low QPS.

\subsection{Explore the Design Space by FPGAs}
\label{sec:FPGA}

Due to the many design options for an IVF-PQ accelerator as introduced above, we leverage the reconfigurability of FPGAs to show the trade-offs between different designs and to compare performance between these designs.

FPGAs are reprogrammable circuits that consist of BRAM and URAM as fast on-chip memory, Flip-Flops (FF) as registers, and Digital Signal Processors (DSP) as computation units. Various FPGA models have different amounts of those resources. 
An FPGA-based accelerator usually consists of a number of processing elements (PEs) to implement functionalities and FIFOs to connect the PEs. For example, a neural network accelerator can allocate the computation workload to multiple matrix multiplication PEs and gather the partial results using another PE, and FIFOs enable the streaming communication between these PEs. 
Developing FPGA accelerators typically requires much more effort than software. Traditionally, FPGAs are developed by hardware description languages (HDL), such as Verilog and VHDL, but recent advances in High-Level Synthesis (HLS) allow programmers to develop the circuit at a higher level using C/C++ or OpenCL~\cite{vivado_hls, intel_opencl_fpga}. FPGAs are now widely available in data centers and in the cloud~\cite{jiang2023data}.

%% file: system_overview.tex
\section{\textit{FANNS} Framework Overview}
\label{sec:system_overview}

\begin{figure*}%[t]
  % full width, can be adjusted
  \centering
  \includegraphics[width=1.0\linewidth]{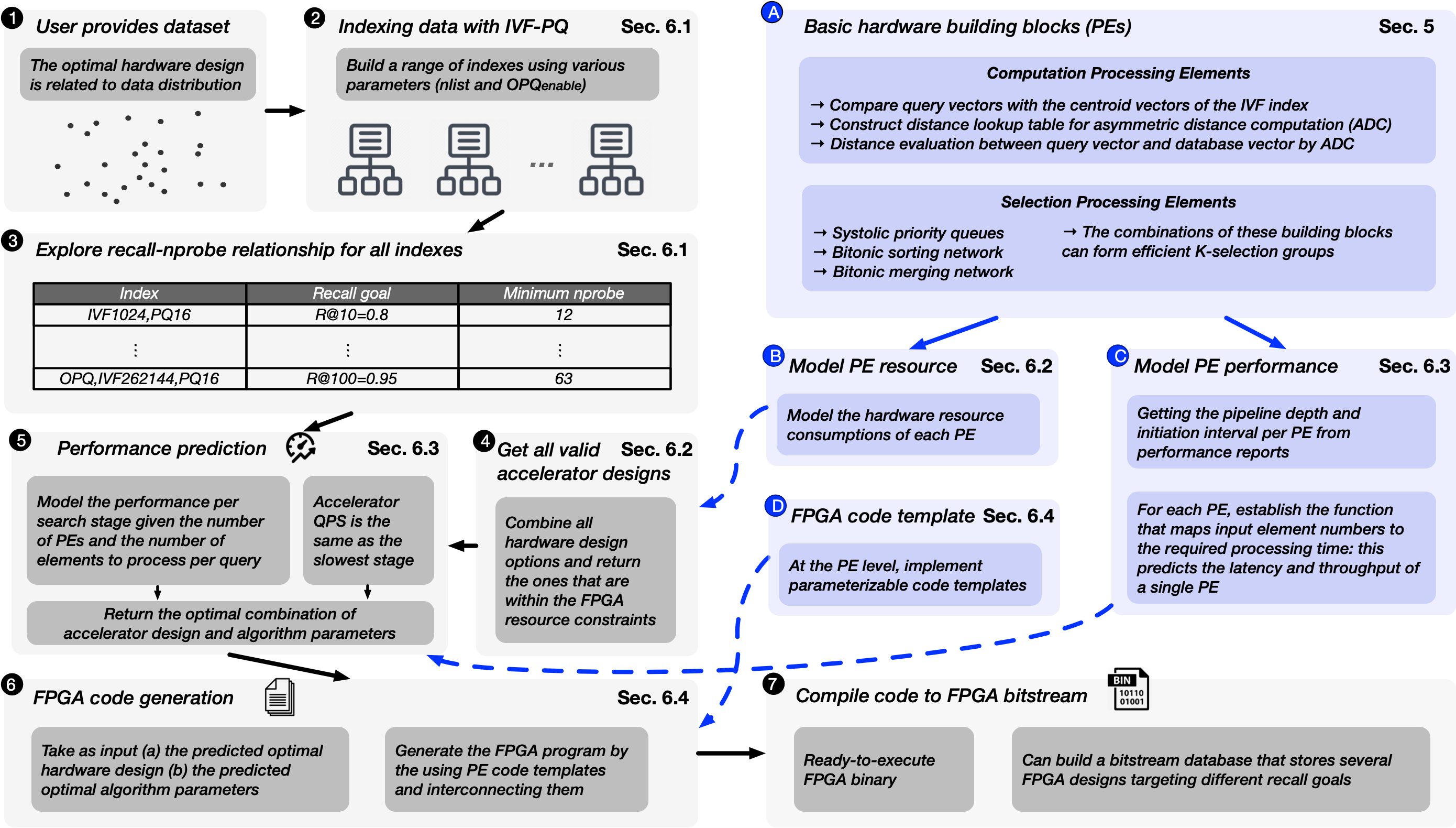}
  \vspace*{-2em}
%   (\ballnumber{1}$\sim$\ballnumber{6}) Seems if I add this, latex compile forever...
  \caption{
  The workflow of \textit{FANNS}. The letter-labeled blue blocks are the framework building blocks independent of user requirements, while the digit-labeled gray blocks are the automatic accelerator generation steps. }
  % \vspace*{-1em}
  \label{fig:workflow}
\end{figure*}

% \begin{figure}%[t]
%   % full width, can be adjusted
%   \centering
%   \includegraphics[width=1.0\linewidth]{fig/workflow-mini.png}
%   % \vspace*{-1.5em}
% %   (\ballnumber{1}$\sim$\ballnumber{6}) Seems if I add this, latex compile forever...
%   \caption{
%   The workflow of \textit{FANNS}. The letter-labeled blue blocks are the framework building blocks independent of user requirements, while the digit-labeled gray blocks are the automatic accelerator generation steps. }
%   % \vspace*{-1.5em}
%   \label{fig:workflow}
% \end{figure}

\begin{figure*}%[t]
  % full width, can be adjusted
  \centering
  \includegraphics[width=1.0\linewidth]{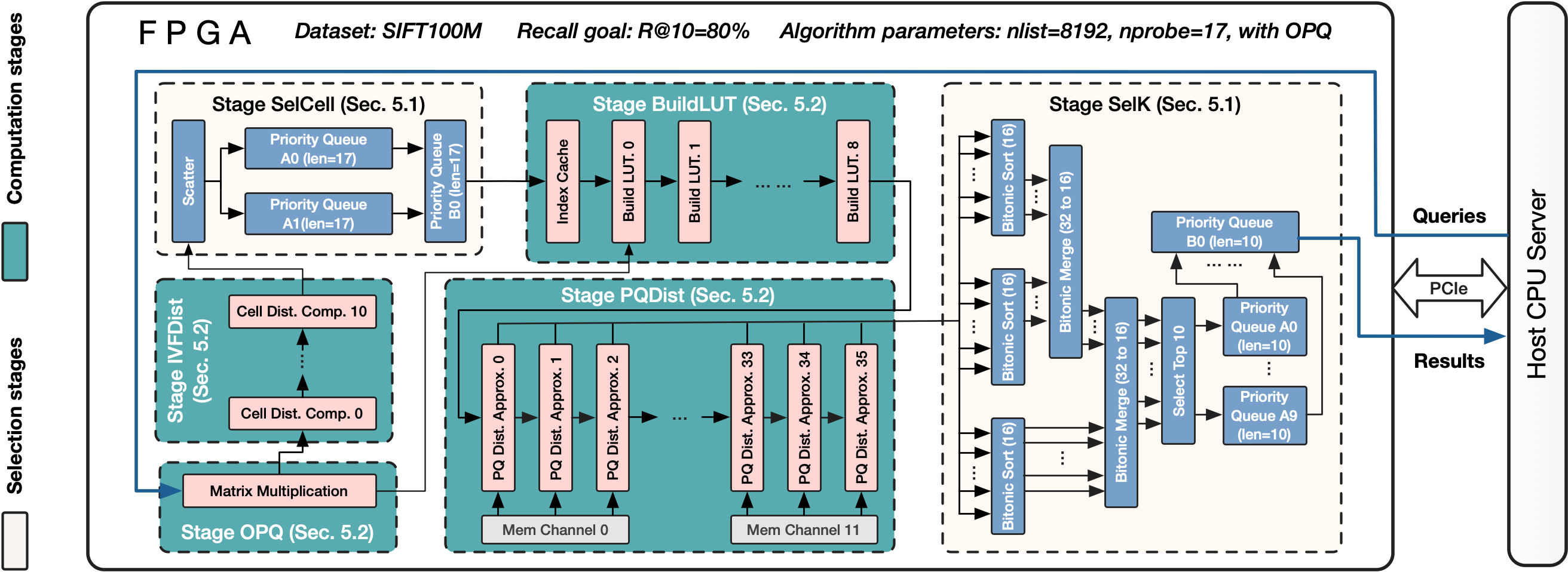}
  \vspace{-2em}
%   (\ballnumber{1}$\sim$\ballnumber{6}) Seems if I add this, latex compile forever...
  \caption{An example accelerator design generated by \textit{FANNS}.}
  \vspace{-1em}
  \label{fig:fpga-sample-design}
\end{figure*}

% Due to the huge design space as we explain in Section~\ref{sec:design_space}, it is impossible to design an one-size-fit-all accelerator that always achieves the best performance given arbitrary recall goals and datasets. 
% For each recall goal on a dataset, there should be an unique accelerator design that achieves the best performance (we mainly target throughput in queries per second (QPS) because FPGAs inherently guarantees low latency).
% However, we first need an effective approach to identify the optimal accelerator design.
% In this paper, we target to find the unique optimal solution per user requirement and automate the hardware design flow.

% One potential solution is to leverage the reconfigurability of FPGA to design different accelerators to fulfill various user requirements. 
% However, that requires: we need a solution to make sure the designed hardware
% Besides, re-designing the accelerator means re-writing the FPGA code, which is time-consumping if user changes recall goals frequently.

We present \textit{FANNS} (FPGA-accelerated Approximate Nearest Neighbor Search), an end-to-end vector search framework by hardware-algorithm co-design. 
\textit{FANNS} targets the deployment scenario where the user (deployer) has a target recall goal (e.g., for all queries, achieve 80\% recall for top 10 results on average) on a given dataset and a given hardware device. 
In this case, \textit{FANNS} can automatically figure out the optimal combination of algorithm parameters and hardware design, and generate the specialized FPGA accelerator for the combination.
\textit{FANNS} also supports scale-out by instantiating a hardware network stack~\cite{100gbps} in the accelerator. 

% With \textit{FANNS}, a user can simply provide the dataset and the recall requirement, while the framework will generate the optimal accelerator and select the appropriate set of parameters. 
% If the user prefers to prepare several levels of recall to handle the fluctuating query arrival pattern (recall-QPS trade-off), the user can generate several accelerators using \textit{FANNS} and reprogram the FPGA when needed. Here we assume the user will not reconfigure the FPGA at the per-query granurality, i.e., setting different recall goals for several queries arrived simultaneously, because reprogramming the FPGA can take up to a second. 

Figure~\ref{fig:workflow} overviews the \textit{FANNS} workflow. 

\textbf{Framework building blocks~(\blueballnumber{A}$\sim$~\blueballnumber{D}).}
To build an IVF-PQ accelerator, we first build a set of PEs for all six search stages~\blueballnumber{A}. 
These building blocks are independent to user requirements.
We design multiple PEs per stage when there are several valid microarchitecture solutions. 
Given the designed PEs, we naturally know their hardware resource consumptions~\blueballnumber{B}.
We can model the PE performance in both latency and throughput~\blueballnumber{C}: knowing the pipeline depth and initiation interval per PE, one can establish the relationship between the number of input elements to process and the respective time consumption in clock cycles. 
Finishing the PE design step, we also have a set of PE code templates~\blueballnumber{D}. 

\textbf{Automatic accelerator generation workflow~(\ballnumber{1}$\sim$\ballnumber{7}).}
The gray blocks in Figure~\ref{fig:workflow}~presents the automatic workflow that customizes the hardware per user recall requirement.
The inputs of the framework are the user-provided dataset and recall goal~\ballnumber{1}. 
Given the dataset, \textit{FANNS} trains a number of indexes using a range of parameters~\ballnumber{2}. 
% One of the trained index will be selected to deploy on FPGA later on. 
Then, for each index, \textit{FANNS} evaluates the relationship between \textit{nprobe} and recall~\ballnumber{3}. 
On the other hand, \textit{FANNS} returns all valid hardware designs whose resource consumption is under the constraint of the given FPGA device~\ballnumber{4}.
Subsequently, \textit{FANNS} uses a performance model to predict the optimal combination of parameter setting and accelerator design~\ballnumber{5}. The performance model takes two input sources: (a) the set of all possible accelerator designs by combining different hardware-level options summarized in Table~\ref{tab:design_space} and (b) the minimal \textit{nprobe} per index given the recall requirement. For each combination of the hardware-level and parameter-level choices, \textit{FANNS} performance model can predict QPS based on per-PE performance. 
Given the predicted optimal design, \textit{FANNS} code generator outputs the ready-to-compile FPGA code by instantiating the respective PEs and interconnecting them~\ballnumber{6}. 
Finally, the FPGA code is compiled to bitstream (FPGA executable)~\ballnumber{7}. 
Table~\ref{tab:time} breaks down the time consumption of the \textit{FANNS} workflow.

\begin{table}[t]
\begin{small}
  \begin{center}
    \caption{Time consumption of the \textit{FANNS} workflow.}
    \label{tab:time}
    \vspace{-1em}
    \begin{tabular}{L{14em} L{14em}} % <-- Alignments: 1st column left, 2nd middle and 3rd right, with vertical lines in between
      \toprule
      \multicolumn{1}{c}{Step} & \multicolumn{1}{c}{Time consumption} \\
      \midrule
      Build Indexes & Several hours per index. \\ % for billion-scale datasets.  \\
      Get recall-nprobe relationship & Up to minutes per index. \\
      Predict optimal design & Up to one hour per recall goal. \\
      FPGA code generation & Within seconds. \\
      FPGA bitstream generation  & Around ten hours per design. \\
      \bottomrule
    \end{tabular}
  \end{center}
   \vspace*{-1.5em} % to shrink gap between figures
\end{small}
\end{table}

\textbf{Framework deployment.} Given its ability to optimize accelerator performance based on specific datasets and recall objectives, FANNS is well-suited for integration into production vector search systems. 
Such systems often manage dynamic datasets, subject to regular insertions and deletions. This is accomplished through the maintenance of a primary IVF-PQ index for a specific dataset snapshot, an incremental (usually graph-based) index for new vectors added since the last snapshot, and a bitmap to track deleted vectors. These two indexes are periodically merged, e.g., once a week, into a new primary index~\cite{adb-v}. In this scenario, FANNS targets optimizing performance for the main index, thus also periodically redesigning accelerators for the new dataset snapshot and, if applicable, the new recall goal. When building the accelerator for the new snapshot, the existing accelerator and the CPU's incremental index continue to process queries. As such, the time taken to build the new accelerator is effectively concealed by the ongoing operation of the older system, barring the initial build. This setup also allows FANNS to always target a static dataset snapshot. The algorithm explorer, therefore, does not need to handle any shifts in data distribution, allowing accurate performance modeling.

\textbf{Example FPGA design.}
Figure~\ref{fig:fpga-sample-design} shows a generated accelerator targeting R@10=80\% on the SIFT100M dataset. 
In this single-accelerator-search scenario, the FPGA communicates with the host CPU through PCIe to receive query requests and return results.
\textit{FANNS} processes queries in a deeply pipelined fashion: there can be multiple queries on the fly in different stages in order to maximize throughput. 
Each stage of processing is accomplished by a collection of PEs.
The arrows connecting those PEs are FIFOs: a PE loads values from the input FIFO(s), processes the inputs, and pushes the results to the output FIFO(s). 
A stage can contain homogeneous PEs such as Stage IVFDist or heterogeneous PEs such as Stage SelK which involves sorting networks, merging networks, and priority queues. 
The PE numbers are typically irregular (11 in Stage IVFDist, 9 in Stage BuildLUT, etc.) as they are calculated by the performance model, unlike being restricted to the exponential of two which human designers favor. 
We will specify the hardware design per search stage in the following section. 
% The reason we use multiple PEs in parallel (e.g., in stage 2) rather than a single large PE is that the hardware compiler (Vivado) performs best when mapping a small piece of digital logic to a small hardware region — using a single huge PE will typically lead to compilation (place and route) failures. 
% The hardware contains two stages, and the PE numbers are irregular, unlike human-designed accelerator. We will specify the hardware design per search stage in the following section. 

%% file: hardware_design.tex
\section{Hardware Processing Elements}
\vspace{-2em}\hspace{24em}\includegraphics[height=0.08\linewidth]{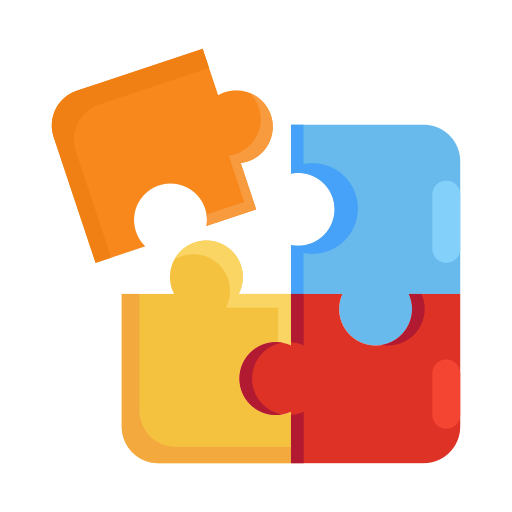}\hfill
\label{sec:hardware_design_space}

We present the accelerator hardware processing elements and the design choices. We group the six search stages into selection stages and computation stages to explain related concepts together. 

\subsection{Designs for the Selection Stages}

Two stages need selection functionality. Stage SelCells selects the closest Voronoi cells to the query vector, given a set of input distances. Stage SelK collects the $K$ smallest distances between the query vector and database vectors, given the many approximated distances output by Stage PQDist every clock cycle. Since there can be multiple PEs producing inputs to the two stages, the selection hardware should support multiple input streams. 

\subsubsection{$K$-Selection Primitives}
% \hfill \\
Bitonic sort networks and systolic priority queues are the building blocks for $K$-selection. 

% Bitonic sort is a parallel sorting algorithm that takes several input elements in parallel, performs a certain series of compare-swap operations, and outputs the sorted array. Bitonic sort exhibits high sorting throughput, and its parallelism aligns very well with FPGAs~\cite{batcher1968sorting, mueller2012sorting, papaphilippou2018flims, song2016parallel, salamat2021nascent, papaphilippou2020adaptable, matai2016resolve}.
% Systolic priority queue~\cite{leiserson1979systolic, huang2014scalable} is a register array interconnected by compare-swap units, allowing 1 replace operation (if the incoming number is smaller than the current root node in the queue, dequeue the root and enqueue the incoming number) every two clock cycles. 
% In the first cycle, the leftmost node is replaced with a new item, and all the even entries in the array are swapped with the odd entries. In the second cycle, all the odd entries are swapped with the even entries.
% % In an odd cycle, the $i$th ($i$ is odd) element in the queue is compare-swapped with the $(i+1)$st element; in an even cycle, the $(i+1)$st element is compare-swapped with the $(i+2)$nd element.
% During this process, the smaller elements are gradually swapped to the rightmost side of the queue. 

% \begin{figure}[t]
%   % full width, can be adjusted
%   \centering
%   \includegraphics[width=0.95\linewidth]{fig/Bitonic-sort-8.png}
%   \vspace{-1em}
%   \caption{A hardware bitonic sorting network.}
%   \vspace{-1em}
%   \label{fig:bitonic-sort}
% \end{figure}

\textit{Bitonic Sort.} Bitonic sort is a parallel sorting algorithm that takes several input elements in parallel, performs a certain series of compare-swap operations, and outputs the sorted array. Bitonic sort exhibits high sorting throughput, and its parallelism aligns very well with FPGAs~\cite{batcher1968sorting, mueller2012sorting, papaphilippou2018flims, song2016parallel, salamat2021nascent, papaphilippou2020adaptable}.
As a result, the latency of sorting an array is $\sum_{i=1}^{log_{2}l}i=\frac{log_{2}l * (1 + log_{2}l)}{2}$ clock cycles where $l$ is the width of the sorting network.
% matai2016resolve}.

% Figure~\ref{fig:bitonic-sort} shows a bitonic sorting network that takes eight unordered elements as the input (values before step 1.1) and outputs the sorted numbers (values after step 3.3).
% The sorting network divides a sequence of numbers into multiple sub-sequences and merges two consecutive sub-sequences as a longer sorted sub-sequence until the entire sequence is sorted.
% As a result, the latency of sorting an array is $\sum_{i=1}^{log_{2}l}i=\frac{log_{2}l * (1 + log_{2}l)}{2}$ clock cycles where $l$ is the width of the sorting network.

% Figure~\ref{fig:bitonic-sort} shows a bitonic sorting network that takes eight unordered elements as the input (values before step 1.1) and outputs the sorted numbers (values after step 3.3).
% The sorting network devides a sequence of numbers into multiple sub-sequences and merge two consecutive sub-sequences as a longer sorted sub-sequence until the entire sequence is sorted.
% As a result, the latency of sorting an array is $\sum_{i=1}^{log_{2}l}i=\frac{log_{2}l * (1 + log_{2}l)}{2}$ clock cycles where $l$ is the width of the sorting network.

% \vspace{0.5em}
% \ul{Bitonic Partial Merge.} As shown in the green region of Figure~\ref{fig:bitonic-sort}, a bitonic partial merger takes two sorted arrays of length $l$ as inputs and outputs the sorted $l$ smallest (or largest) numbers. The hardware implementation of a bitonic partial merger is a subset of a single step of a bitonic sorter. A pipelined hardware design allows $l$ output elements per cycle.

\begin{figure}[t]
  % full width, can be adjusted
  \centering
  \includegraphics[width=0.95\linewidth]{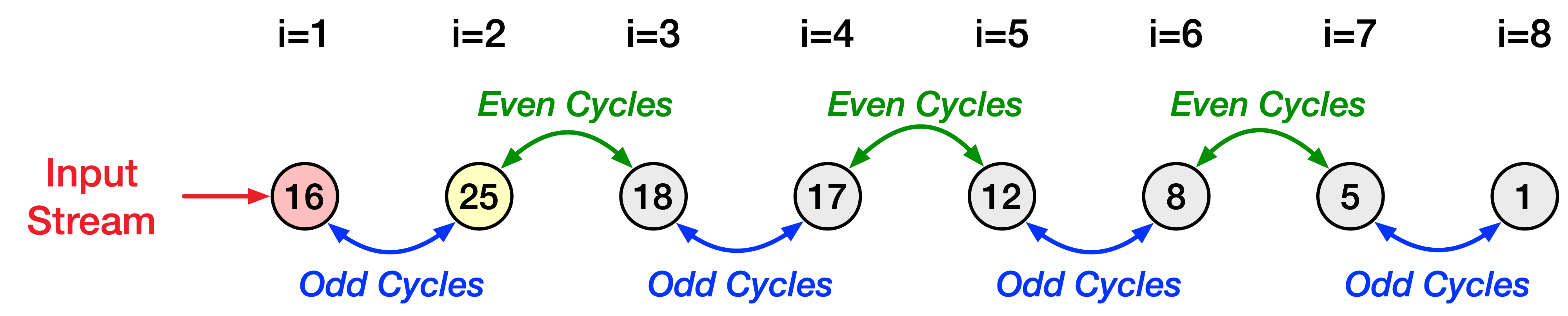}
  \vspace{-1em} 
  \caption{A hardware systolic priority queue.}
  \vspace{-1.5em}
%   \vspace*{-3mm}
  \label{fig:priority-queue}
\end{figure}

\textit{Systolic Priority Queue.} 
While software-based priority queues support enqueue, dequeue, and replace operations, we only need the replace operation: if the input is smaller than the current root, dequeue the root and enqueue the input.
% Systolic priority queue~\cite{leiserson1979systolic, huang2014scalable} is a register array interconnected by compare-swap units, allowing 1 replace operation every two clock cycles. 
Figure~\ref{fig:priority-queue} shows in the implemented systolic priority queue~\cite{huang2014scalable, leiserson1979systolic} that supports such minimal required functionality while consuming the least hardware resources. 
% Systolic priority queue~\cite{huang2014scalable, leiserson1979systolic} supports such minimal required functionality while consuming the least hardware resources. 
It is a register array interconnected by compare swap units, supporting one replace operation every two clock cycles. 
In the first cycle, the leftmost node is replaced with a new item, and all the even entries in the array are swapped with the odd entries. 
In the second cycle, all the odd entries are swapped with the even entries.
% In an odd cycle, the $i$th ($i$ is odd) element in the queue is compare-swapped with the $(i+1)$st element. Note that the root element, in this case, is the minimum of the old root and the incoming number. 
% In an even cycle, the $(i+1)$st element is compare-swapped with the $(i+2)$nd element.
During this process, the smallest elements are gradually swapped to one side of the queue.

\subsubsection{$K$-Selection Microarchitecture Design}
% \hfill \\
Parallel $K$-selection collects the $s$ smallest numbers per query ($s=nprobe$ in Stage SelCells; $s=K$ in Stage SelK) out of $z$ input streams given that each stream produces $v$ values per query. We propose two design options for this task with different trade-offs:

% \ul{Design 1: hierarchical priority queue (HPQ).} We propose 
% \begin{wrapfigure}{l}{0.5\linewidth}
%   % full width, can be adjusted
%   \vspace{-1em}
%   \includegraphics[width=\linewidth]{fig/2-level-priority-queue.png}
%   \vspace{-2em}
%   \caption{A two-level hierarchical priority queue.}
%   \vspace*{-1em}
%   \label{fig:hierarchical-priority-queue}
% \end{wrapfigure}
% % We propose HPQ as a straightforward way for parallel selection. 
\textbf{Option 1: hierarchical priority queue (HPQ).} We propose HPQ as a straightforward way for parallel selection. 
The first level of HPQ contains $z$ queues to collect $s$ elements from each stream.
The second level takes the $zs$ elements collected in the first level and selects the $s$ results.
% When there are many queues in the first level, the only queue in the second level can become the performance bottleneck. In this case, HPQ should involve more levels of queues to maximize the overall selection performance. 
The HPQ allows $z/2$ input elements per cycle since each replace operation in a priority queue requires two cycles. As a result, if an input stream generates one element per cycle, we should split it into two substreams and match it with two priority queues in the first level.

% The number of queues per level decreases as the level increases.
% The key idea is to hold $s$ selections from each stream in the first level and gather these $zs$ values through the rest levels of priority queues: as the level increases, the selections are narrowed down.
% It has several levels of queues of the same depth of $s$, where the first level consumes the $z$ input streams. This design allows $z/2$ input elements per cycle since each replace operation in a queue requires two cycles. As a result, if a computation PE can generate one distance per cycle, we need to match it with two priority queues in the first level.
% The number of levels depends on $s$, $z$, and $v$.
% For example, when setting the level number to be 2, we must make sure the second level of only one queue will not bound the performance. Considering each queue has the same insertion performance, the insertion number of the level-2 queue ($sz$) should be smaller than the insertion per level-1 queue ($v$). This indicates that when $v$ is large while $s$ and/or $z$ are small, two levels of queues are enough.
% On the other hand, we use multiple levels of queues when $v<sz$ so that the higher levels do not bound the overall performance. 
% As a result, the exact numbers of levels and queues need to be calculated given $s$, $z$, and $v$.
% However, it is worth noticing that the hardware resource consumption can be high if we need a lot of those queues when $z$ is large and/or the level number is large. 

\begin{figure}%[t]
  % full width, can be adjusted
  \centering
  \includegraphics[width=1.0\linewidth]{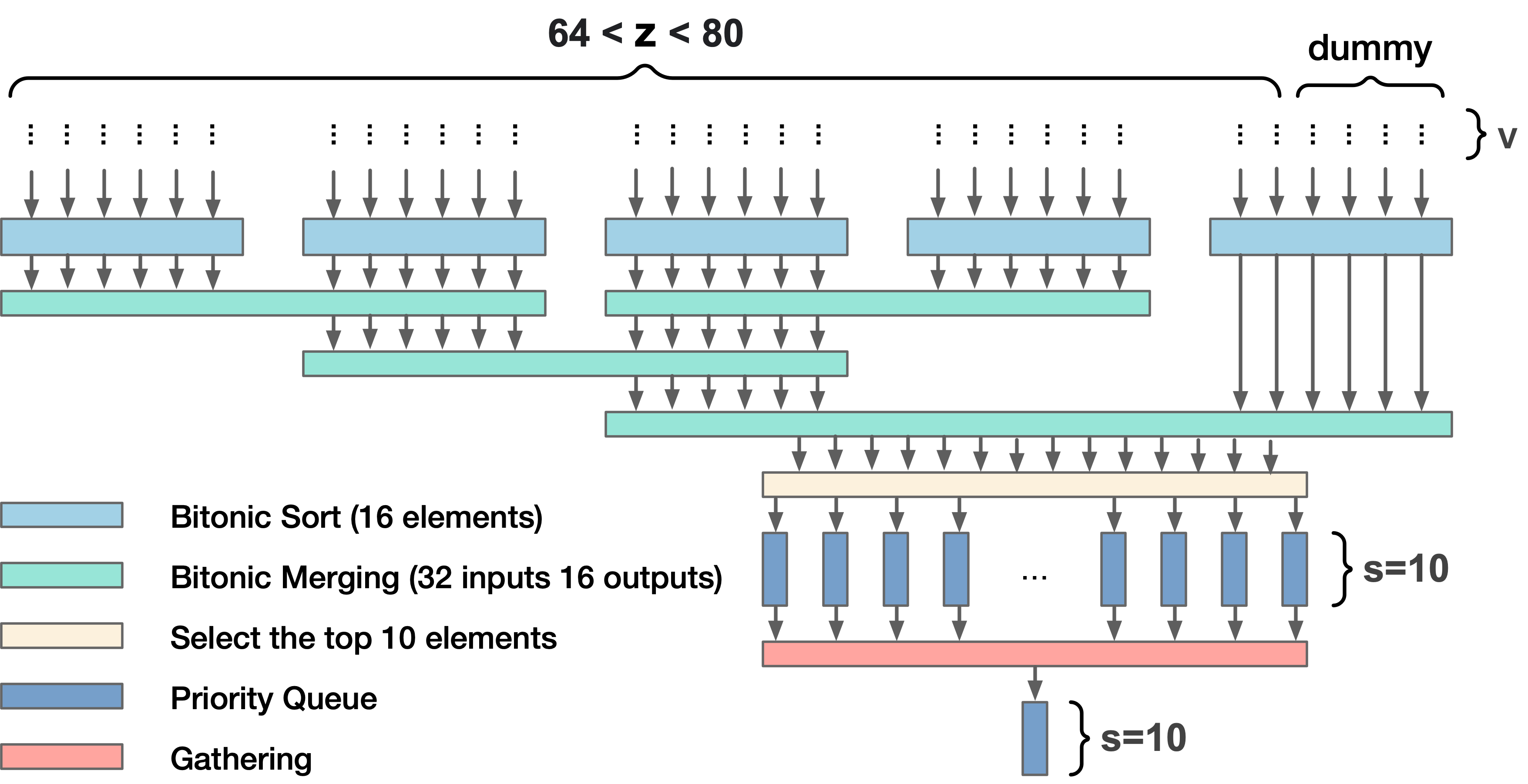}
  \vspace{-2em}
  \caption{An example of hybrid bitonic sorting, merging, and priority queue architecture that selects the top 10 elements out of up to 80 input streams.}
  \vspace{-2em}
%   \vspace*{-3mm}
  \label{fig:hybrid-topk}
\end{figure}

\textbf{Option 2: hybrid sorting, merging, and priority queue group (HSMPQG).} 
The key idea is to collect the $s$ results per clock cycle before inserting them into the priority queues, such that the number of required queues can be significantly reduced.
Figure~\ref{fig:hybrid-topk} shows an example of such design ($64<z\leq80$ and $s=10$). 
The first step is to sort every 16 elements since 16 is the minimum bitonic sort width greater than $s=10$. Handling up to 80 inputs per cycle requires five bitonic sort networks. Some dummy streams are added as the input for the last sorting network. 
The second step is to merge the sorted elements by several bitonic partial mergers. Each bitonic merger outputs the top 16 elements out of the two input sorted arrays. After several merging steps, we have the sorted top 16 elements per cycle.
Afterward, the $s=10$ elements per cycle are picked out of the 16 and inserted into a hierarchical priority queue, which outputs the $s$ results per query.
Note that we can configure the number of bitonic sort and parallel merge networks for different workloads. For example, if $16<z\leq32$, two sorting and one merging modules are required; we will need three sorting and two merging networks when $32<z\leq48$.

% As shown in Figure~\ref{fig:hybrid-topk}, this design targets the case when $v$ is larger than $s$. 
% \textit{Why using 16 as the sorting network size in HSMPQG?} 
% The sorting network size is at least 10 to filter out $K=10$ elements per cycle. 
% Using a small size of 10 means more sorting and merging PEs are instantiated, consuming more total hardware resources than a few larger PEs (due to the higher interface overhead in small PEs).
% However, the hardware compiler struggles to successfully map huge PEs to FPGAs.
% Overall, a sorting network size of 16 is a compromise to achieve good resource efficiency and to guarantee successful compilation.

\textbf{Intuition behind different $K$-selection microarchitecture.} 
The HPQ design suits the situation when the input stream number $z$ is small, because the few priority queues instantiated will not consume many resources. This design is also the only option when $s\geq{z}$, for which the second option cannot filter out unnecessary elements per cycle at all.
The HSMPQG design targets to collect a small result set over many input streams. It could save hardware resources by significantly reducing the number of priority queues compared with the first option. However, the bitonic sorting and merging networks also count for resource consumption, thus the second option is not always better even if $s<z$.
% Note that given a specific workload, the design choice should be made after a careful evaluation of the performance and resource consumption of both designs.
% , rather than directly picking one based on intuition.

% \textbf{TODO: not ony hierachical PQ, but also sorting all results.}

\subsection{Designs for the Computation Stages}

Computation stages include Stage OPQ, Stage IVFDist, Stage BuildLUT, and Stage PQDist.
In this section, we first specify the Stage PQDist PEs to convey the compute PE design principles on FPGAs, and then introduce the PE interconnection topology. 

% Stages OPQ transforms the query vectors with a rotation matrix. 
% Stage IVFDist evaluates the distances between the transformed query vector and all Voronoi cell centroids in the IVF index.
% Stage BuildLUT builds distance lookup tables per Voronoi cell per query. 
% Stage PQDist estimates the distances between query vectors and quantized database vectors.
% The index can be stored either on-chip (SRAM) or off-chip (HBM), while the PQ codes representing database vectors are assigned to be stored in HBM.
% % For Stage PQDist, the number of PEs decides the number of required HBM channels, as each  relationship
% We adopt a 1-D array PE lineup to forward data between PEs within a stage: one PE gets input from the FIFO connected to the previous PE and outputs data to the next PE.
% Such 1-D connection contrasts the one-to-many broadcasting/gathering strategy that results in high fan-in and fan-out issues and eventually leads to failures in placement \& routing (P\&R).
% We follow the high-performance HLS programming guidelines~\cite{de2020transformations} to design each compute PE.

\subsubsection{Stage PQDist.}
% \hfill \\  
As shown in Figure~\ref{fig:fpga-sample-design}, there are many Stage PQDist PEs working in parallel, approximating distances between the query vector and the quantized database vectors.
% In this section, we first introduce how the input is fed into the PEs and then discuss how the PEs handle the distance approximation.

\begin{figure}%[t]
  % full width, can be adjusted
  \centering
  \includegraphics[width=1.0\linewidth]{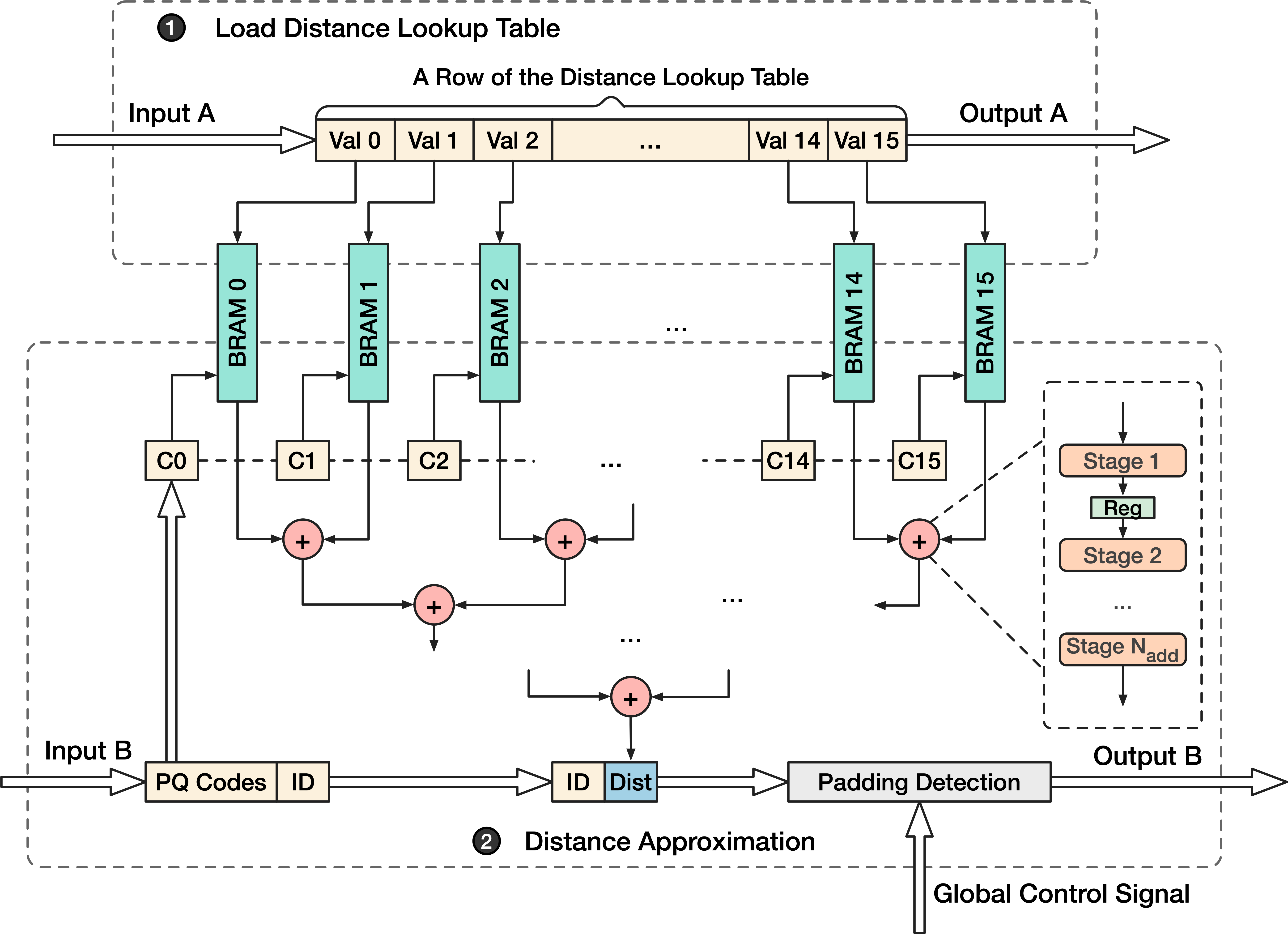}
  \vspace{-2em}
  \caption{The PE hardware design for Stage PQDist.}
  \vspace{-2em}
  \label{fig:PE_PQDist}
\end{figure}

\textbf{PE design.} 
Figure~\ref{fig:PE_PQDist} presents the PE design for decoding 16-byte PQ codes.
The PE takes two inputs: the distance lookup tables produced by Stage BuildLUT and the PQ codes stored in off-chip memory channels. 
For a single query, a PE repeats two major steps $nprobe$ times.
The first step is reading a distance lookup table of size of $km$.
We use BRAM, a type of fast on-chip SRAM, to cache the tables. %Each BRAM slice can store 18Kbit data and can either read or write a 32-bit value per cycle.
In order to provide memory access concurrency for the computing step, we assign $m$ BRAM slices per PE --- each slice stores a column of a table. 
% To minimize the table initialization time, we set the input and output FIFOs for the tables as 512-bit wide (16 BRAM slices can read $16\times32=512$ bits per cycle). 
The second step is approximating the distances between the query vector and the database vectors by asymmetric distance computation.
Each PQ code (1-byte) of a database vector is used as the lookup index for one column of a distance table, and $m$ distances are retrieved from the BRAM slices in parallel per cycle. 
These partial distances are then fed into an add tree that produces the total distance.
In order to maximize the computation throughput, each logical operator (addition) in the tree is composed of several DSPs (for computation) and FFs (as registers), such that the computation is fully pipelined, allowing the add tree to consume $m$ input distances and to output one result per clock cycle.
During the last iteration of scanning a cell, the PE performs padding detection and overwrites the output by a large distance for the padded case. 
The meta-info about padding is streamed into the PE by the accelerator's global controller.

\textbf{PE size.} 
% The second design choice is the PE size --- how large should it be? 
In principle, a PE with more computation logic is usually more efficient in terms of performance delivered per hardware resource unit. This is because each PE has some surrounding logic as the interface to other PEs --- the smaller each PE, the more significant the total overhead. However, it is hard for the FPGA compiler to map a huge chunk of logic to the FPGA successfully~\cite{de2020flexible}.
% (e.g., SIMD width of 128 floating point numbers). 
Thus, we experiment with several PE sizes and select the largest one that can be successfully compiled. 

\subsubsection{PE interconnection Topology.}
\hfill \\
Within a computation stage, we adopt a 1-D array architecture to forward data between the homogeneous PEs.
For example, the second PE consumes the query vector and the results of the first PE, appends its local results, and sends the query vector as well as the aggregated results to the third PE. 
Another design choice, which we do not adopt, is the broadcasting/gather topology.
The advantage of the 1-D array architecture over the broadcasting/gather one is the minimized wire fan-out: too many long wires connected to a single source can lead to placement \& routing failure during FPGA compilation~\cite{de2020transformations}.
For communication between stages and within a selection stage, the FIFO connections are straightforward because there is no input sharing as in computation stages.

%% file: e2e.tex
\section{End-to-End Hardware Generation}
\label{sec:e2e}

This section illustrates the end-to-end accelerator generation flow of \textit{FANNS}, as visualized in Figure~\ref{fig:workflow} (\ballnumber{1}$\sim$\ballnumber{7}). We implement the end-to-end generation flow using a set of Python scripts, while the hardware processing elements are implemented in Vitis HLS.

% \subsection{Implementing Hardware Generation}

% We implement the end-to-end generation flow using a set of python scripts. The python scripts takes as input the accelerator information containing the resource consumption and produce a list of valid accelerators. It also takes as input the performance modeling parameters, thus evaluating the performance of each valid accelerator and save the best combination of algorithm parameter set and the hardware design as a configuration file. The code generation script load the configuration and use the hardware processing element PEs to generate the ready-to-compile hardware.

\vspace{-1em}
\subsection{Explore Algorithm Parameters~\includegraphics[height=0.06\linewidth]{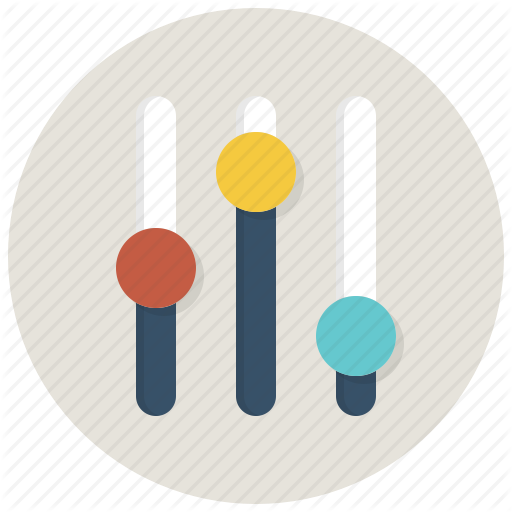}}
\label{sec:design_space_exploration_nprobe}

\textbf{Given a dataset, \textit{FANNS} first captures the relationship between the algorithm parameters and recall, which will be used for accelerator QPS prediction.}
% Since each dataset has a unique data distribution, the first step to build the data-aware accelerator is understanding the relationship between algorithm parameters and recall. 
\textit{FANNS} trains a number of IVF indexes trying different \textit{nlist}~\ballnumber{2}. 
Each index is trained both with and without OPQ. 
Given the user-provided sample query set, \textit{FANNS} evaluates the minimum \textit{nprobe} that can achieve the user-specified recall goal on each index~\ballnumber{3} (e.g., 80\% of average recall for top 10 results). 
The result of this step is a list of index-\textit{nprobe} pairs that serve as the inputs of the FPGA performance model.
% using the 10,000 query vectors in the SIFT dataset.

\vspace{-1em}
\subsection{List Valid Accelerator Designs~\includegraphics[height=0.06\linewidth]{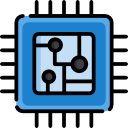}}
\label{sec:resource_model}

\textbf{\textit{FANNS} lists all valid accelerator designs on a given FPGA device by resource consumption modeling}~\ballnumber{4}.
Specifically, \textit{FANNS} combines all hardware choices in Table~\ref{tab:design_space} to form accelerators and returns the valid ones whose consumptions of all types of resources (BRAM, URAM, LUT, FF, and DSP) are under the device constraint. 
Consuming all resources on the FPGA is unrealistic because such designs will fail at the placement \& routing step during FPGA compilation. 
But it is also impossible to predict the maximum resource utilization per design because the EDA algorithms for FPGA compilation are nondeterministic. 
As a result, we set the resource utilization rate as a constant for all accelerators, e.g., a conservative value of 60\% in our experiments.
% The maximum percentage of resource utilization is a constant. 
% The second one is the resource consumption constraint: although large designs can lead to placement \& routing failures during FPGA compilation, we cannot predict the maximum resource utilization per design, which typically ranges from 70\%$\sim$90\%.
% In this paper, we set a conservative resource utilization constraint as 60\%.

\vspace{-1em}
\begin{equation}
\label{eq:resource_model}
% \begin{aligned*}
% \vspace{-1em}
\scalebox{0.8}{
$
\begin{gathered}
% \begin{align*}
    \sum_{i} C_{r}(PE_i) + \sum_{i} C_{r}(FIFO_i) + C_{r}(infra) \leq Constraint_{r} , \\
    \forall r\in\{BRAM, URAM, LUT, FF, DSP\}
% \end{align*}
\end{gathered}
$
} % scalebox
% \end{aligned*}
% \vspace{-1em}
\end{equation}

\textit{FANNS} models an accelerator's resource consumption by summing up several components as in Equation~\ref{eq:resource_model} ($C_r$ denotes the consumption of resource $r$).
The first part is of all the PEs. The consumption of a PE is known once we finish designing and testing the PE. For priority queues of variable lengths, we employ a linear consumption estimation model since the numbers of registers and compare-swap units in a priority queue are linear to the queue length. 
The second part is the FIFOs connecting PEs, which can be modeled by measuring the consumption of a single FIFO and counting the required FIFO numbers.
% The third source is the global controller orchestrating the PEs in all stages. It cannot be measured before having the entire accelerator. Fortunately, there is no heavy computation within the controller, thus we can assume its consumption to be negligible.
The final component is the infrastructure surrounding the accelerator kernel, such as the memory controller, which consumes constant resources. 
% We consider an accelerator design valid if the sum of these resource consumptions is lower than the predefined threshold.  

\vspace{-1em}
\subsection{Model Accelerator Performance~\includegraphics[height=0.06\linewidth]{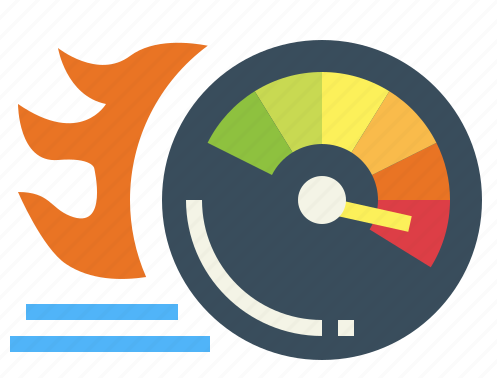}}
\label{sec:design_space_exploration_perf_modeling}

\textbf{The \textit{FANNS} performance model predicts the QPS of all combinations of algorithm parameters and accelerator designs, then returns the optimal one.}
Given the large design space, it is unrealistic to evaluate QPS by compiling all accelerators and testing all parameters.
Thus, we need an effective performance model to predict the QPS per combination~\ballnumber{5}. By using the following modeling strategy, \textit{FANNS} can evaluate all (millions of) combinations given a recall requirement within an hour. We now introduce the model in a top-down manner. 

\textit{Model the performance of an accelerator.} As the six search stages of IVF-PQ are pipelined, the throughput of the entire accelerator is that of the slowest stage, as in Equation~\ref{eq:performance_all}. 
% Thus, we need to identify and maximize the performance of the bottleneck stage. 
% This is because our accelerator processes queries in a deeply pipelined manner, i.e., there will be one query in each stage at peak time, thus the $QPS_{all}$ is determined by the slowest stage. 
\begin{equation}
\label{eq:performance_all}
% \begin{aligned*}
\scalebox{0.95}{$
\begin{gathered}
% \begin{align*}
    QPS_{accelerator} = min(QPS_{s})\mathrm{, where~} s\in \mathrm{\{Stages}\}
% \end{align*}
\end{gathered}
$}
\end{equation}

\textit{Model the performance of a search stage.} 
A search stage typically consists of multiple PEs functioning concurrently.
If these PEs share the same amount of workload, the time consumption per query of the stage is the same as the time consumption for a single PE to handle its own workload.
% and the workloads are dispatched to each PE as balanced as possible.
If the workloads are imbalanced per PE, the performance of the stage is decided by its slowest PE. 
% Given the number of PEs in a search stage and the performance model of a single PE, we can model the performance of the search stage. 
% If the workloads are perfectly balanced across the PEs, the throughput of a stage is linear to the number of PEs. 
% Otherwise, the performance is decided by the slowest PE in the stage.

% \begin{equation}
% \label{eq:performance_all}
% % \begin{aligned*}
% \begin{gathered}
% % \begin{align*}
%     QPS_{stage} = min(QPS_{p})\mathrm{, where~} p\in \mathrm{\{PEs}\}
% % \end{align*}
% \end{gathered}
% \end{equation}

\textit{Model the performance of a PE.} 
Inspired by de Fine Licht et al.~\cite{de2020transformations}, we estimate the throughput of a single PE by predicting the number of clock cycles it takes to process a query ($CC$). 
For a single query, we suppose that the PE processes $N$ input elements. 
The pipeline initiation interval is $II$, which represents the number of cycles that must pass before a new input can be accepted into the pipeline. 
The pipeline has a latency $L$, which is the number of cycles it consumes for an input to propagate through the pipeline and arrive at the exit.
$L$ and $II$ are known constants after implementing the hardware. $N$ can be either a constant or a variable. For example, after deciding the algorithm parameters and the accelerator design, the number of distances evaluated per PE in Stage IVFDist is a constant (\textit{N=nlist/PENum}). By contrast, the number of PQ codes scanned in Stage PQDist differs for every query due to the imbalanced number of codes per cell. In this case, we estimate $N$ by taking the expected scanned entries per query (assume the query vector distribution is identical to the database vectors, such that cells containing more vectors are more likely to be scanned). 
Given the numbers of $L$, $II$ and $N$, we can estimate the consumed clock cycles as $CC=L+(N-1)*II$. The QPS of the PE can then be derived by Equation~\ref{eq:performance_pe} where \textit{freq} is the accelerator frequency. 
Similar to predicting the maximum resource utilization rate, it is impossible to know the operational frequency of an accelerator before compilation.
Thus, we assume the frequency to be a constant for all accelerators. % e.g., 140 MHz in our experiments. 
% The first one is clock frequency, because it is impossible to predict the accurate accelerator frequency before compile time.
% Based on our experience on the U280 FPGA, we set the default frequency as 140MHz in our experiments. 
% We assume the clock frequency to be a known constant because it is impossible to predict frequency accurately at design time. 

\vspace{-1em}
\begin{equation}
\label{eq:performance_pe}
% \begin{aligned*}
\begin{gathered}
% \begin{align*}
    QPS_{PE} = freq/(L+(N-1)*II)
    % CC=L+(N-1)*II\\
    % QPS_{PE} = freq/CC
% \end{align*}
\end{gathered}
\end{equation}

% \ul{Example.} 

% \begin{equation}
% \label{eq:performance_resource_model_object}
% % \begin{aligned*}
% \begin{gathered}
% % \begin{align*}
%     QPS_{accelerator} = min(QPS_{s})\mathrm{, where~} s\in \mathrm{\{Stages}\}\\
%     QPS_{i}(design_{j}, param_{k}) = \frac{freq}{CC_i(design_{j}, param_{k})}\mathrm{, where~}\\ 
%     j\in \mathrm{\{Hardware~designs\}}~\mathrm{and}~k\in \mathrm{\{Parameter~settings\}}~\mathrm{s.t.} \\
%     Consum_{k}(design_j) \leq ConsumLimit_k, \forall k\in{\{\mathrm{Resource\ types\}}}.
% % \end{align*}
% \end{gathered}
% % \end{aligned*}
% \end{equation}

% The model contains a couple of constant coefficients decided by the user. 
% The first one is clock frequency, because it is impossible to predict the accurate accelerator frequency before compile time.
% Based on our experience on the U280 FPGA, we set the default frequency as 140MHz in our experiments. 
% The second one is the resource consumption constraint: although large designs can lead to placement \& routing failures during FPGA compilation, we cannot predict the maximum resource utilization per design, which typically ranges from 70\%$\sim$90\%.
% In this paper, we set a conservative resource utilization constraint as 60\%.

\vspace{-1em}
\subsection{Generate FPGA Programs~\includegraphics[height=0.06\linewidth]{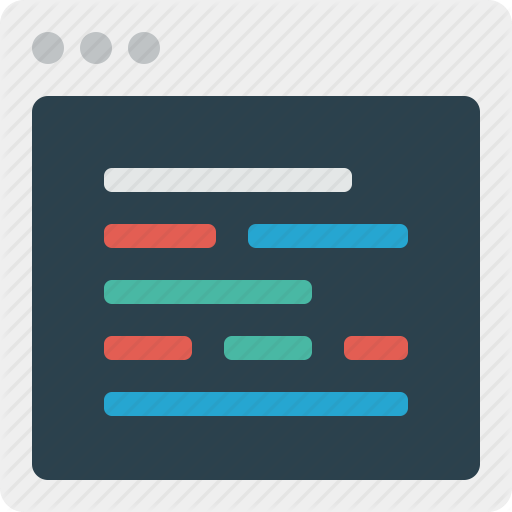}}
\label{sec:design_space_exploration_codegen}

{\textit{FANNS} code generator takes as inputs the optimal combination of parameter setting and hardware design and produces the ready-to-compile FPGA code.}
% Given the implemented PEs introduced in Section~\ref{sec:hardware_design_space}, we already have the respective code templates. 
Refering to the inputs, the code generator instantiates the given numbers of PEs using the PE code templates, the respective on-chip memory for index caching, the FIFOs interconnecting the aforementioned components, and the off-chip memory interfaces between the accelerator kernel and the FPGA shell~\ballnumber{6}.
Since the code generation step does not involve complex logic, it only consumes seconds to return the FPGA program, which will be further compiled to the bitstream~\ballnumber{7}.

%% file: evaluation.tex
\section{Evaluation}
\label{sec:evaluation}

This section shows the effectiveness and necessity of algorithm-hardware co-design to achieve the optimal vector search performance on FPGAs. 
% Such results indicate that future ASIC designs should avoid the shifting bottlenecks by only targeting large-scale search and by cooperating with a general-purpose processor. We present the key results in this section and leave more details (such as the specification of \textit{FANNS}-generated designs) in Appendix~\ref{sec:appendix_generated_hardware} and \ref{sec:appendix_gpu_performance}.
We also integrate the FPGAs with network stacks to show their great scalability.

\subsection{Experimental Setup}
\label{sec:target_platform}

\textbf{Baseline.}
We compare \textit{FANNS} with CPU, GPU, and FPGA baselines. 
The CPU and GPU baselines run Faiss (version 1.7.0), a popular ANN library developed by Meta known for its efficient IVF-PQ implementation.
The FPGA baseline uses the same set of hardware building blocks as \textit{FANNS} but without being parameter-aware.
% Apart from CPU and GPU baselines, we also compare \textit{FANNS} with an FPGA baseline as we will introduce later, which does not involve the automatic design space exploration flow. 
% For FPGA, the QPS remains the same with or without batching because the proposed design naturally operates on per-query granularity.

\textbf{Hardware Setup.}
We choose CPUs, GPUs, and FPGAs that are manufactured in the generation of technology. 
We use an m5.4xlarge CPU server on AWS, which contains 16 vCPUs of 16 vCPUs of Intel(R) Xeon(R) Platinum 8259CL @ 2.50GHz (Cascade Lake, 14nm technology) and 64 GB of DDR4 memory.
We use NVIDIA V100 GPUs (CUDA version 11.3) manufactured by the TSMC 12 nm FFN (FinFET NVIDIA) technology (5,120 CUDA cores; 32 GB HBM).
We use Xilinx Alveo U55c FPGA fabricated with TSMC's 16nm process. It contains 1.3M LUTs, 9K DSPs, 40MB on-chip memory, and 16 GB HBM. We develop the accelerators using Vitis HLS 2022.1. 
%  HBM (8GB, 32 channels, up to 400GB/s)

\textbf{Benchmark.}
We evaluate \textit{FANNS} on standard and representative vector ANN benchmarks: the SIFT and Deep datasets.
% The SIFT dataset~\cite{SIFT, jegou2011searching} contains several size scales ranging from 1 million to 1 billion 128-dimensional vectors. 
The SIFT dataset contains 128-dimensional vectors, while the Deep dataset consists of 96-dimensional vectors.
For both datasets, we adopt the 100-million-vector size scale as they can fit into the FPGA memory after product quantization. 
Both datasets contain 10,000 query vectors and respective ground truths of nearest neighbor search for recall evaluation.
% The Deep dataset contains 96-dimensional vectors, and we choose the 100-million subset (Deep100M).
% For all experiments (CPU, GPU, FPGA), the entire dataset is loaded in memory (DDR4 for the CPU, HBM2 for GPUs and FPGAs) without any disk I/Os. 
We set various recall goals on each dataset. 
As recalls are related to $K$ (the more results returned, the more likely they overlap with true nearest neighbors) and the data distribution, we set one recall goal per K per dataset, i.e., R@1=30\%, R@10=80\%, and R@100=95\% on the SIFT dataset and R@1=30\%, R@10=70\%, and R@100=95\% on the Deep dataset.
% On SIFT100M dataset, the maximum achievable R@1 using with 16-byte quantization is less than 30\%, while R@100 can surpass 95\%.

\textbf{Parameters.} 
% \hfill \\
% \textit{Index options.} 
We explore a range of algorithm parameters and set a couple of constant factors for \textit{FANNS} performance model.
On the algorithm side, we trained a range of indexes with different numbers of Voronoi cells (\textit{nlist} ranges from $2^{10}$ to $2^{18}$ ) for each dataset, so as to achieve the best QPS for not only FPGA but the CPU and GPU baselines. 
Per \textit{nlist}, we trained two indexes with and without OPQ to compare the performance.
% \textit{Compression rate.}
We quantize the vectors to 16-byte PQ codes ($m=16$) for all indexes and all types of hardware.
The primary consideration is to fit the dataset within FPGA's device memory while achieving high recall. 
On the \textit{FANNS} side, we set the maximum FPGA resource utilization rate as 60\% to avoid placement \& routing failures. We also set the target accelerator frequency as 140MHz based on our design experience with the U55c FPGA device.

\subsection{\textit{FANNS}-Generated Accelerators}

This section presents the \textit{FANNS} generated accelerators. We show that the optimal designs shift with parameter settings. We then present the fully customized accelerator designs under target recall and compare them against the parameter-independent FPGA baseline design.

\begin{figure*}[t]
  % \vspace*{-5mm} % to shrink gap between figures
  \centering
  
  \begin{subfigure}[b]{0.33\linewidth}
    \includegraphics[width=\linewidth, height=8em]{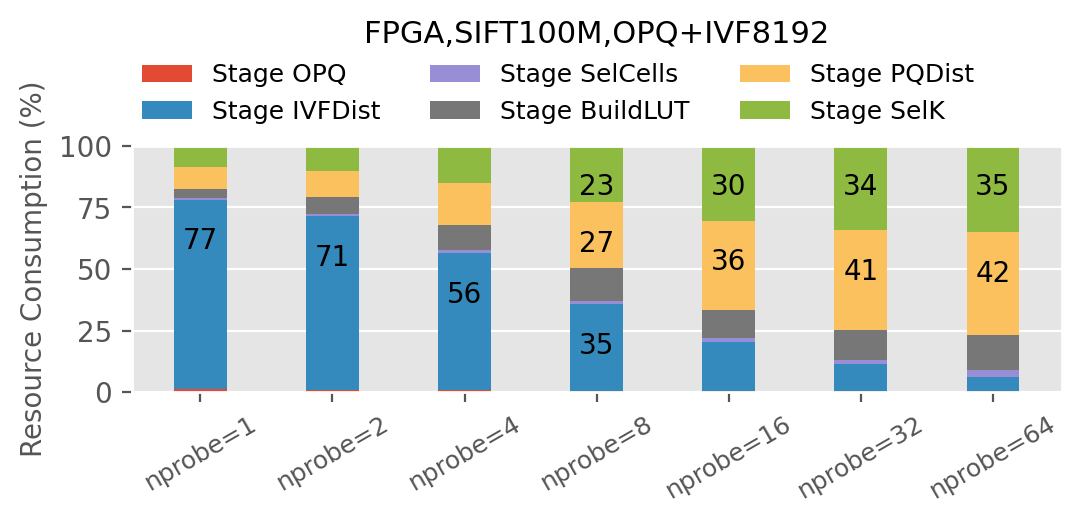}
    % \caption{Tasty coffee.}
  \end{subfigure}
  \hfill
  \begin{subfigure}[b]{0.33\linewidth}
    \includegraphics[width=\linewidth, height=8em]{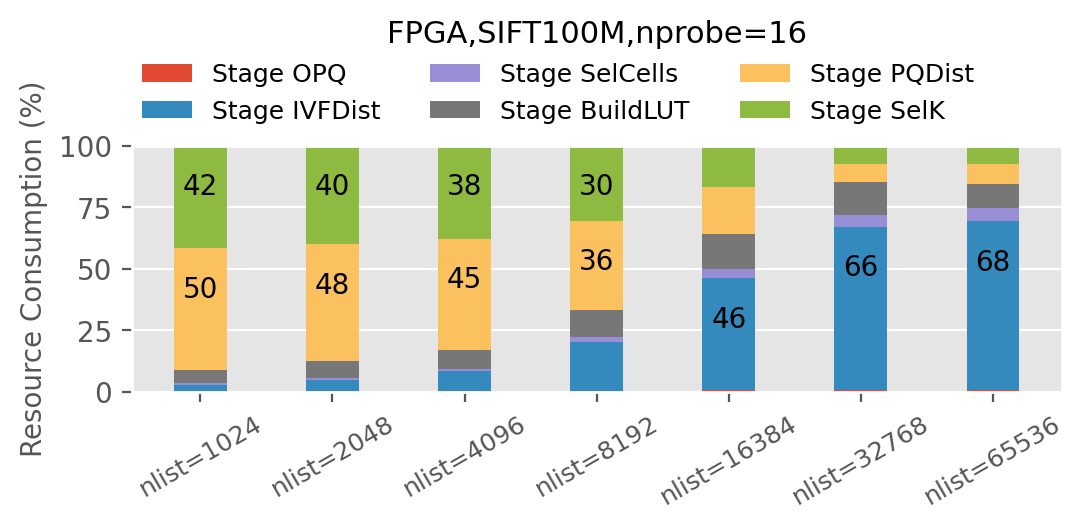}
    % \caption{Tasty coffee.}
  \end{subfigure}
  \hfill
  \begin{subfigure}[b]{0.33\linewidth}
    \includegraphics[width=\linewidth, height=8em]{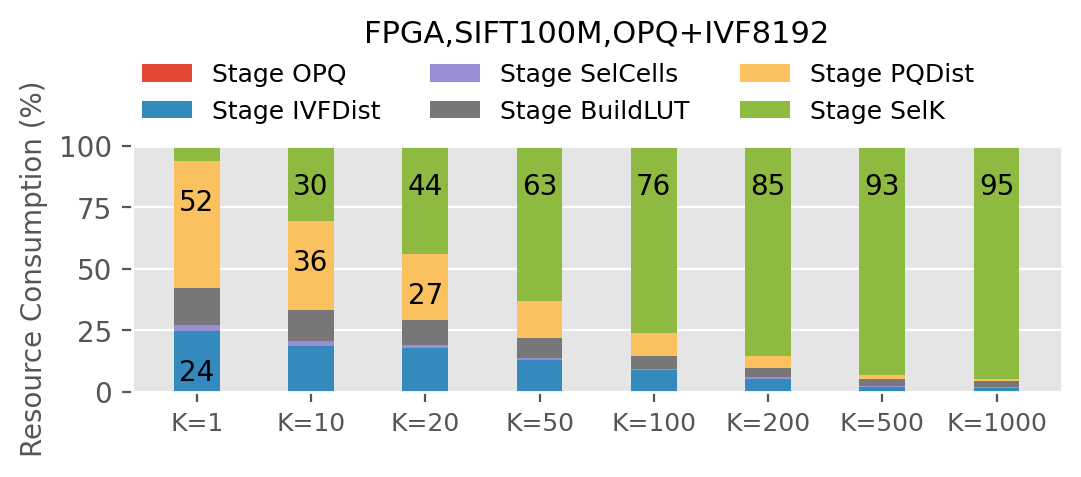}
    % \caption{Tasty coffee.}
  \end{subfigure}

  \vspace{-1em}
  \caption{The optimal FPGA designs shift with algorithm parameters (left: \textit{nprobe}; middle: \textit{nlist}; right: $K$.)}
  \label{fig:FPGA_design_shift}
\vspace*{-0.5em}
  % \vspace*{-5mm} % to shrink gap between figures
\end{figure*}

\subsubsection{The Effect of Algorithm Parameters on Hardware Designs}
\hfill \\
\textbf{Optimal accelerator designs shift significantly with algorithm parameters, as shown in Figure}{~\ref{fig:FPGA_design_shift}}. In this experiment, we assign the parameters to the \textit{FANNS} performance model, which predicts the optimal hardware design under the parameter constraints. To draw high-level conclusions from the designs, we only visualize the resource consumption ratio per search stage, omitting the microarchitecture-level choices.
First, we observe the effect of \textit{nprobe} on the designs. 
% We use the index per hardware that can achieve the highest performance of R@100=95\%. For FPGAs, we use R@10=80\% because large $K$ can result in most resources used in instantiating priority queues.
As the number of PQ codes to scan increases, more hardware resources are dedicated to Stage PQDist and Stage SelK, while most of the resources are allocated to Stage IVFDist when $nprobe$ is small.
Then, we fix \textit{nprobe} and observe the designs when shifting \textit{nlist}. As \textit{nlist} raises, more PEs are instantiated for Stage IVFDist so as to improve the performance of centroid distance computation. 
Finally, as $K$ increases, the resources spent on Stage SelK surge because the resource consumption of hardware priority queues is linear to the queue length $K$.

\subsubsection{The Optimal Accelerator Designs of Given Recall Goals}
\hfill \\
Table~\ref{tab:final_designs}~summarizes the \textit{FANNS}-generated designs per recall and compares them with the baseline designs. It shows that: 

\textbf{First, \textit{FANNS} picks different parameters per recall.} For the three recall requirements, \textit{FANNS} adopts different indexes and $nprobe$ to maximze the performance.

\textbf{Second, \textit{FANNS} generates different hardware designs for each recall requirement.} Stage SelK, for example, applies the two different microarchitecture designs (HPQ and HSMPQG) and invests different amounts of hardware resources (2.9\%$\sim$31.7\% LUT) for the three recall requirements. Even for the stages using the same microarchitecture, e.g., Stage IVFDist, the PE numbers of these accelerators can also be different.

\subsubsection{Parameter-independent Accelerator Designs}
\hfill \\
{We design a set of parameter-independent ANNS accelerators that can serve queries on arbitrary indexes as the FPGA baseline. }
% These accelerators have a fixed ratio of hardware resource consumption per search stage.
% Such accelerators share the same microarchitecture primitives as $\textit{FANNS}, but the overall accelerator designs are fixed.
% The ratio of hardware resource consumption is fixed, similar to the Q100 database accelerator~\cite{wu2014q100}.
% We compare the designs with the data-dependent designs that we will introduce later.
% The fixed design approach is based on two observations: first, it is hard to determine the optimal percentage of resources allocated to a stage, because the performance bottleneck depends on algorithms settings; second, the resource consumption of priority queues in Stage SelK is proportional to $K$, thus instantiating long queues (e.g., length of 100) for queries of small $K$ (e.g., 10) wastes resources.
We design three parameter-independent accelerators for different $K$ requirements (1, 10, 100) as shown in Table~\ref{tab:final_designs}. 
Each accelerator roughly balances resource consumption across stages such that the accelerator should perform well on a wide range of algorithm settings. 
Saying this, we do not simply allocate 1/6 resources to each of the six stages due to the following facts. 
First, the number of PEs between Stage PQDist and Stage SelK should be proportional, as more distance computation PEs should be matched with more priority queues to balance the performance between the two stages.
Second, Stage OPQ performs a lightweight vector-matrix multiplication that consumes few resources.
% We thus group the six stages into two stage groups and balance consumption between them: Stage Group A includes Stage OPQ, Stage IVFDist, Stage SelCells, and Stage BuildLUT, while Stage Group B includes Stage PQDist and Stage SelK. 
% Within Stage Group A, all stages but Stage OPQ consume roughly the same amount of resources.

% We design three parameter-independent accelerators for different $K$ requirements (1, 10, 100). 
% Each accelerator roughly balances resource consumption across stages such that the accelerator should perform well on a wide range of algorithm settings. 
% Saying this, we do not simply allocate 1/6 resources to each of the six stages due to the following facts. 
% First, the number of PEs between Stage PQDist and Stage SelK should be proportional, as more distance computation PEs should be matched with more priority queues to balance the performance between the two stages.
% Second, Stage OPQ performs a lightweight vector-matrix multiplication that consumes few resources.
% We thus group the six stages into two stage groups and balance consumption between them: Stage Group A includes Stage OPQ, Stage IVFDist, Stage SelCells, and Stage BuildLUT, while Stage Group B includes Stage PQDist and Stage SelK. 
% Within Stage Group A, all stages but Stage OPQ consume roughly the same amount of resources.
% We present the resulting baseline designs in Table~\ref{tab:final_designs}.

\begin{table*}
%   \ra{1.2}
  \caption{
  Comparison between human-crafted design and \textit{FANNS}-generated designs (for the SIFT100M dataset), including index selection, architectural design, resource consumption (LUT), and predicted QPS.}
  \vspace{-1em}
  \label{tab:final_designs}
% \begin{center}
% \begin{small}
\scalebox{0.57}{ % scale down the table to 0.7x size
\begin{tabular}{
@{} l l c % 1~3 
l c % 4~5
l l c % 6~8 Stage OPQ
l l l c % 9~12 IVFDist
l l l c %  13~16 SelCells
l l l c % 17~19 BuildLUT
l l c % 20~22 PQDist
l l l c % 23~26 SelK
l @{}} % 25
% @{} L{7em}  L{5em} M{0em} % 1~3 
% L{3em} M{0em} % 4~5
% L{3em} L{2em} M{0em} % 6~8 Stage OPQ
% L{3em} L{4.5em} L{2em} M{0em} % 9~12 IVFDist
% L{3em} L{3em} L{2em}  M{0em} %  13~16 SelCells
% L{3em} L{4.5em} L{2em} M{0em} % 17~19 BuildLUT
% L{3em} L{2em} M{0em} % 20~22 PQDist
% L{3em} L{3em} L{2em} M{0em} % 23~26 SelK
% L{4.5em} @{}} % 25
    \toprule
    \phantom{} & \multirow{2}{2em}{Index} & \phantom{} & 
    \multirow{2}{2em}{nprobe} & \phantom{} & 
    \multicolumn{2}{c}{Stage OPQ} & \phantom{} 
    & \multicolumn{3}{c}{Stage IVFDist} & \phantom{} 
    & \multicolumn{3}{c}{Stage SelCells} & \phantom{} 
    & \multicolumn{3}{c}{Stage BuildLUT} & \phantom{} 
    & \multicolumn{2}{c}{Stage PQDist} & \phantom{} 
    & \multicolumn{3}{c}{Stage SelK} & \phantom{} 
    & \multirow{2}{4.5em}{Pred. QPS (140 MHz)} \\
    % \cmidrule{0-4}  \cmidrule{6-12}  
    \cmidrule{6-8} 
    \cmidrule{9-11} 
    \cmidrule{13-15} 
    \cmidrule{17-19} 
    \cmidrule{21-22}
    \cmidrule{24-26}  
    & & & 
    & & 
    \#PE & LUT.(\%) &&
    \#PE & Index store & LUT.(\%) &&
    Arch. & \#InStream & LUT.(\%) &&
    \#PE & Index store & LUT.(\%) &&
    \#PE & LUT.(\%) && 
    Arch. & \#InStream & LUT.(\%) && 
     \\
    \midrule 
    
    K=1 (Baseline) & N/A && 
    N/A && 
    1 & 0.2 &&
    10 & HBM & 6.9 && 
    HPQ & 2 & 6.4 && 
    5 & HBM & 6.9 && 
    36 & 15.2 && 
    HPQ & 72 & 1.8 && 
    N/A  \\
    
    K=10 (Baseline) & N/A &&  
    N/A && 
    1 & 0.2 &&
    10 & HBM & 6.9 &&
    HPQ & 2 & 6.4 &&
    4 & HBM & 6.3 &&
    16 & 6.7 &&
    HPQ & 32 & 5.7 && 
    N/A \\
    
    K=100 (Baseline) & N/A && 
    N/A && 
    1 & 0.2 &&
    10 & HBM & 6.9 &&
    HPQ & 2 & 6.4 &&
    4 & HBM & 6.3 &&
    4 & 1.7 &&
    HPQ & 8 & 15.0 &&
    N/A \\
    
    \midrule 
    % $\mathrm{nlist=4096,nprobe=5}$
    K=1 (\textit{FANNS})  & IVF4096 &&  
    5 && 
    0 & 0 &&
    16 & on-chip & 11.0 &&
    HPQ & 2 & 0.3 &&
    5 & on-chip & 2.6 &&
    57 & 24.0 &&
    HPQ & 114 & 2.9 &&
    31,876 \\
    
    % $\mathrm{OPQ,nlist}=8192\mathrm{,nprobe}=17$
    K=10 (\textit{FANNS})  & OPQ+IVF8192 && 
    17 && 
    1 & 0.2 &&
    11 & on-chip & 7.6 &&
    HPQ & 2 & 0.9 &&
    9 & on-chip &5.2 &&
    36 & 15.2 &&
    HSMPQG & 36 & 12.7 &&
    11,098 \\
    
    % $\mathrm{OPQ,nlist}=16384\mathrm{,nprobe}=33$
    K=100 (\textit{FANNS}) & OPQ+IVF16384 && 
    33 && 
    1 & 0.2 &&
    8 & on-chip & 5.5 &&
    HPQ & 1 & 0.6 &&
    5 & on-chip & 3.6 &&
    9 & 3.8 &&
    HPQ & 18 & 31.7 &&
    3,818 \\
    
    \bottomrule%\hline
  \end{tabular}
  } % scalebox
\end{table*}
\vspace{-1em}

\subsection{Performance Comparison}
\label{sec:overall_performance}

% -- SIFT100M, K 1 --recall_goal 0.3

% Speedup over CPU best baseline: 1.07 x
% Speedup over FPGA best baseline: 1.33 x
% Speedup over CPU worst baseline: 19.12 x
% Speedup over FPGA worst baseline: 20.79 x
% CPU: $1.1\sim19.1$
% FPGA: $1.3\sim20.8$

% --SIFT100M, K 10 --recall_goal 0.8

% Speedup over CPU best baseline: 1.33 x
% Speedup over FPGA best baseline: 1.72 x
% Speedup over CPU worst baseline: 25.84 x
% Speedup over FPGA worst baseline: 8.28 x
% CPU: $1.3\sim25.8$
% FPGA: $1.7\sim8.3$

% --SIFT100M, K 100 --recall_goal 0.95 

% Speedup over CPU best baseline: 0.78 x
% Speedup over FPGA best baseline: 1.44 x
% Speedup over CPU worst baseline: 7.09 x
% Speedup over FPGA worst baseline: 7.89 x
% CPU: $0.8\sim7.1$
% FPGA: $1.4\sim7.9$

% -- Deep100M, K 1 --recall_goal 0.3

% Speedup over CPU best baseline: 1.05 x
% Speedup over FPGA best baseline: 1.26 x
% Speedup over CPU worst baseline: 37.24 x
% Speedup over FPGA worst baseline: 22.99 x

% CPU: $1.1\sim37.2$
% FPGA: $1.3\sim23.0$

% -- Deep100M, K 10 --recall_goal 0.7

% Speedup over CPU best baseline: 1.20 x
% Speedup over FPGA best baseline: 1.77 x
% Speedup over CPU worst baseline: 27.15 x
% Speedup over FPGA worst baseline: 14.76 x

% CPU: $1.2\sim27.2$
% FPGA: $1.8\sim14.8$

% -- Deep100M, K 100 --recall_goal 0.95

% Speedup over CPU best baseline: 0.78 x
% Speedup over FPGA best baseline: 1.48 x
% Speedup over CPU worst baseline: 10.74 x
% Speedup over FPGA worst baseline: 11.95 x

% CPU: $0.8\sim10.7$
% FPGA: $1.5\sim12.0$

% All three cases:
% CPU: $0.8\sim37.2$ (all) $0.8\sim1.3$ (best)
% FPGA: $1.3\sim23.0$ (all)  $1.3\sim1.8$ (best)

\begin{figure*}%[h!]
  % \vspace*{-5mm} % to shrink gap between figures
  \centering
  
  \begin{subfigure}[b]{0.33\linewidth}
    \includegraphics[width=\linewidth]{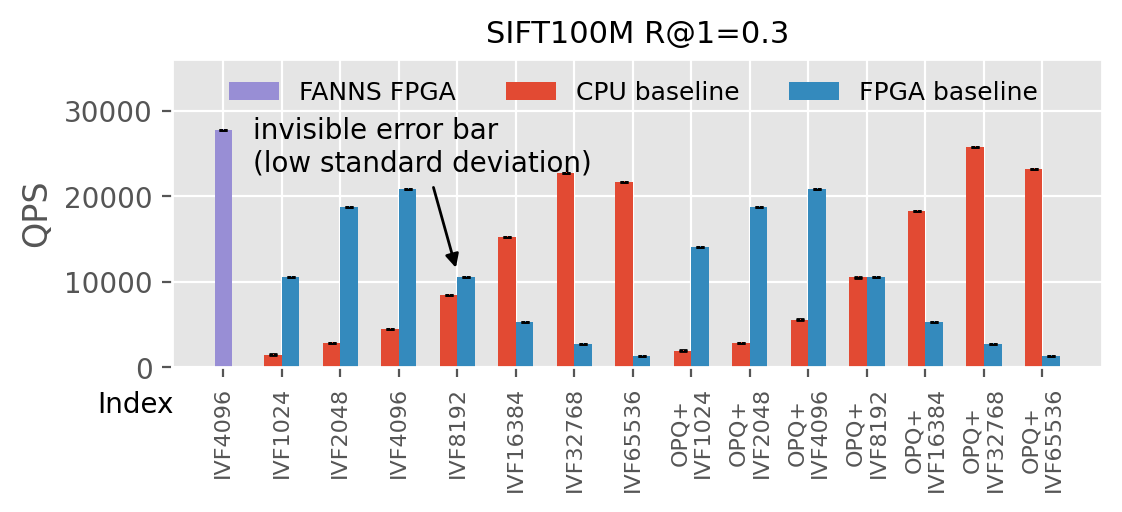}
     % \caption{Coffee.}
  \end{subfigure}
  \begin{subfigure}[b]{0.33\linewidth}
    \includegraphics[width=\linewidth]{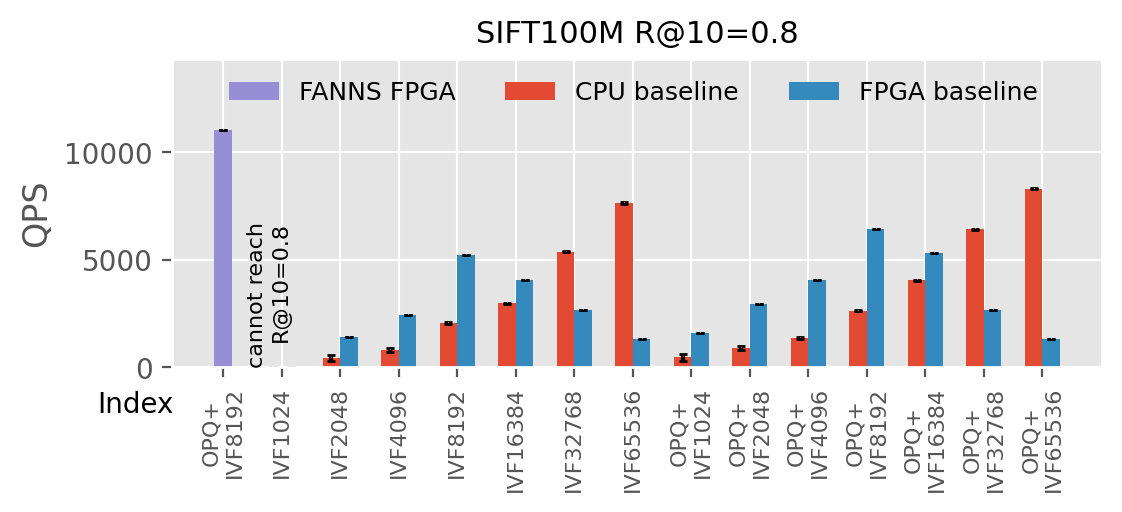}
    % \caption{More coffee.}
  \end{subfigure}
  \begin{subfigure}[b]{0.33\linewidth}
    \includegraphics[width=\linewidth]{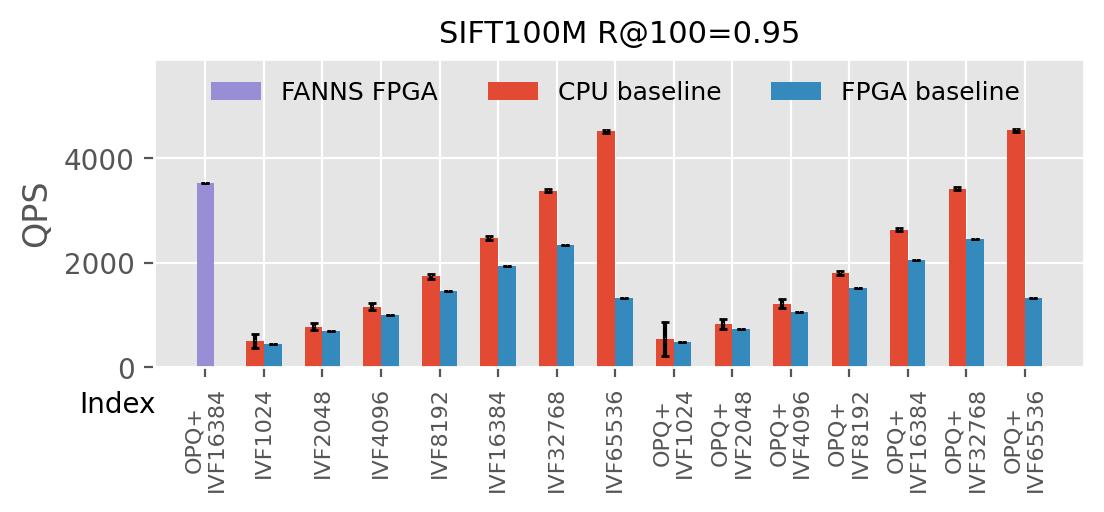}
    % \caption{Tasty coffee.}
  \end{subfigure}
  
  \begin{subfigure}[b]{0.33\linewidth}
    \includegraphics[width=\linewidth]{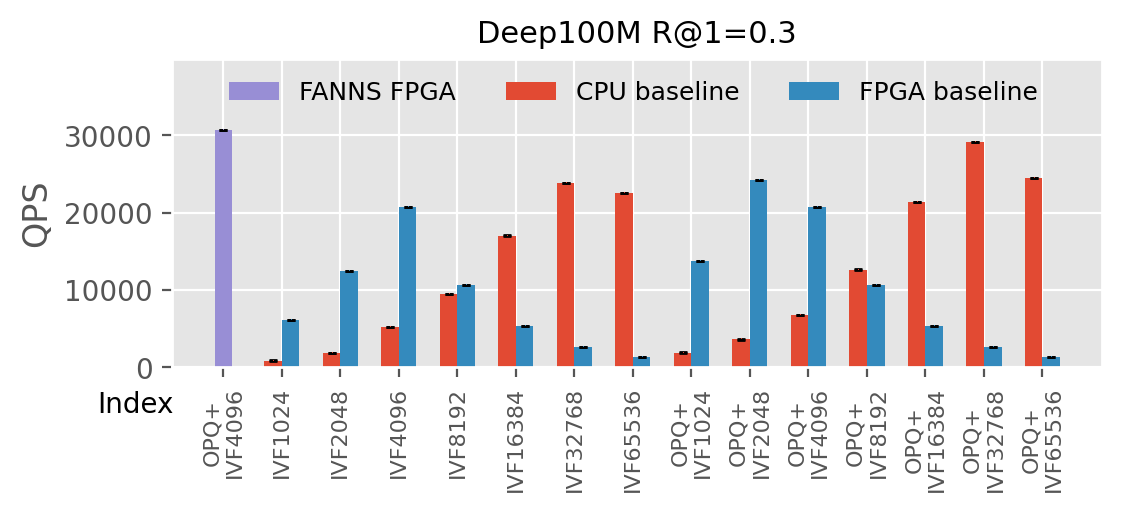}
     % \caption{Coffee.}
  \end{subfigure}
  \begin{subfigure}[b]{0.33\linewidth}
    \includegraphics[width=\linewidth]{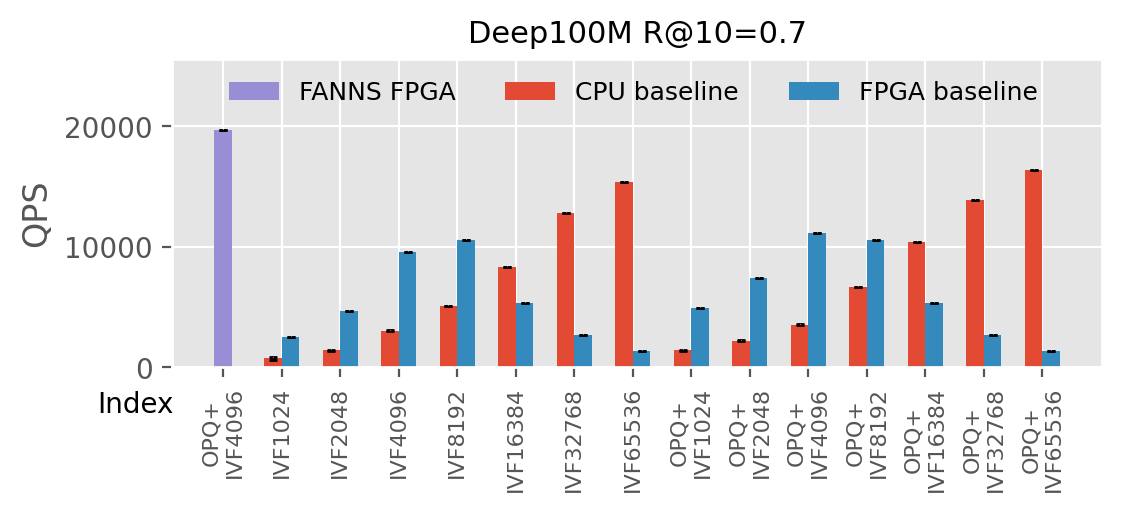}
    % \caption{More coffee.}
  \end{subfigure}
  \begin{subfigure}[b]{0.33\linewidth}
    \includegraphics[width=\linewidth]{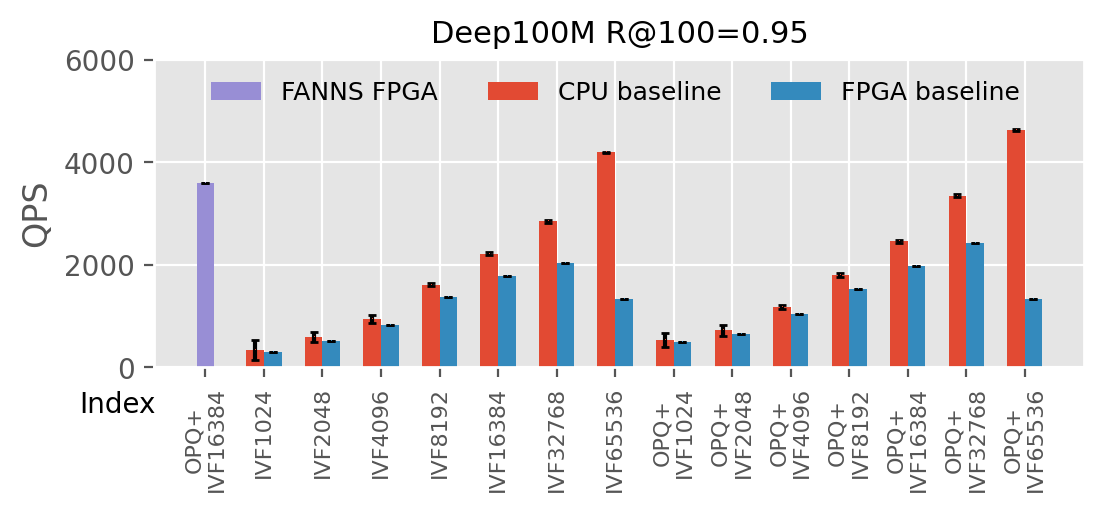}
    % \caption{Tasty coffee.}
  \end{subfigure}
  
  \vspace{-1em} 
  \caption{The throughput comparison between \textit{FANNS}-generated accelerators and CPU/FPGA baselines on the SIFT dataset (first row) and the Deep dataset (second row) under various recall requirements (three columns). }
  \label{fig:fpga_cpu_throughput}
  % \vspace*{-5mm} % to shrink gap between figures
\vspace{-1em}
\end{figure*}

\begin{figure}%[h!]
  % \vspace*{-5mm} % to shrink gap between figures
  \centering

  \begin{subfigure}[b]{0.495\linewidth}
    \includegraphics[width=\linewidth, height=8em]{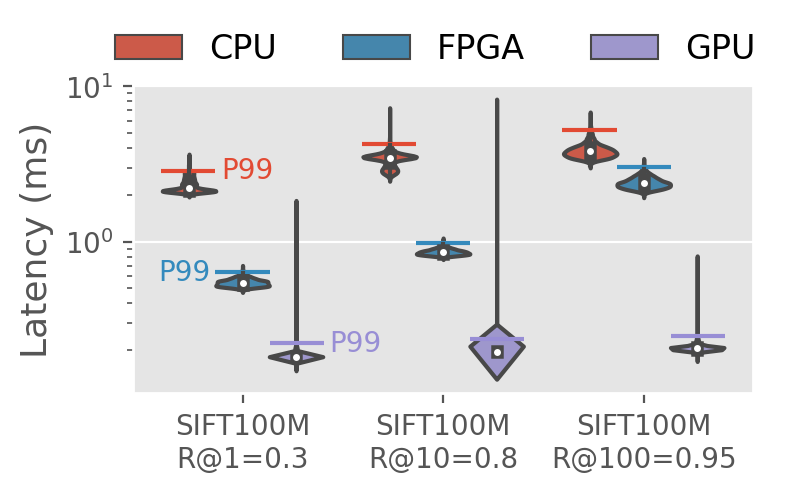}
    % \caption{Tasty coffee.}
  \end{subfigure}
  \hfill
  \begin{subfigure}[b]{0.495\linewidth}
    \includegraphics[width=\linewidth, height=8em]{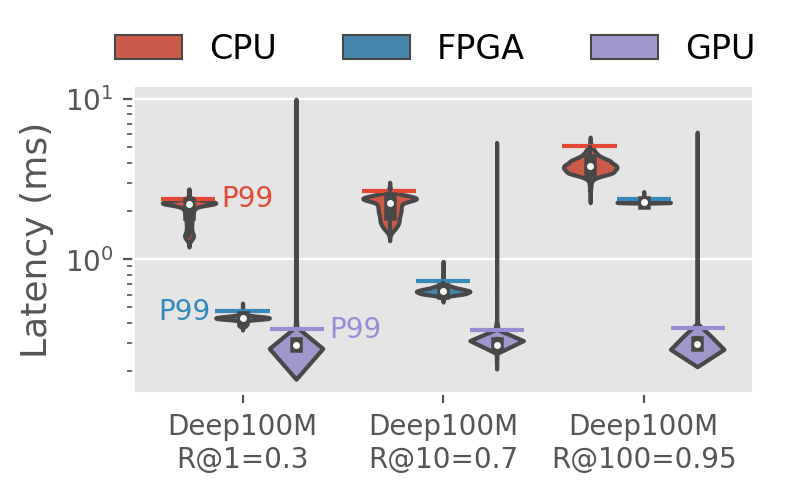}
    % \caption{Tasty coffee.}
  \end{subfigure}

  \vspace{-1em}
  \caption{Latency of single-node CPU, GPU, and FPGA.}
  \vspace{-1em}
  \label{fig:fpga_cpu_latency}
  % \vspace*{-1em} % to shrink gap between figures
\end{figure}

\subsubsection{Offline Batch Processing}
\hfill \\
We first compare the throughput (QPS) between \textit{FANNS} and the CPU/FPGA baselines in Figure~\ref{fig:fpga_cpu_throughput}.
The throughput experiments have no latency constraints, thus allowing query batching (size = 10K) to report the highest QPS.
\textit{FANNS} reports $1.3\sim23.0\times$ QPS as the baseline FPGA designs and $0.8\sim37.2\times$ as the CPU. 
As FPGAs have two orders of magnitude lower flop/s than GPUs, GPU still achieves significantly higher QPS than FPGAs ($5.3\sim22.0\times$), although the FPGAs show comparable latency and better scalability, as we will present later.
Several observations from the throughput experiments include:

\textbf{First, customizing the FPGA per use case is essential to maximize performance.} Although we have done our best to design the parameter-independent FPGA baseline, the \textit{FANNS}-generated accelerators are customized for a target recall requirement on a given dataset, thus showing significant QPS improvements and latency reductions compared with the baseline designs. 
% Even compared with the best-case parameter-independent FPGA performance, our solution achieves 1.33$\sim$1.72$\times$ speedup.

% R@1 27709/31876=86.9%
% R@10 11035/11098=99.4%
% R@100 3519/3818=92.2%
\textbf{Second, the performance model can effectively predict the accelerator performance.} 
By comparing the actual FPGA performance in Figure~\ref{fig:fpga_cpu_throughput} and the \textit{FANNS}-predicted performance in all experiments, we find the actual QPS can reach 86.9\%$\sim$99.4\% of the predicted performance.
In the case when the generated accelerators can achieve the target frequency, the actual performance is virtually the same as the predicted one.
When the target frequency cannot be met due to the nondeterministic FPGA placement and routing algorithm, the achieved performance drops almost proportionally with the frequency. 
% For R@10=80\%, the actual performance is virtually the same as the predicted one (99.4\%). 
% The actual performance of R@1=30\% and R@100=95\% is slightly lower than the prediction due to a couple of reasons. First, we find that Vivado did not achieve the target frequency of 140MHz, based on which the performance is predicted. The final frequencies are 126 MHz and 139 MHz, respectively. Second, PEs are connected by FIFOs, and full FIFOs at runtime can stall the pipeline. 
% For R@10=80\%, the actual performance is virtually the same as the predicted one (99.4\%). 
% The actual performance of R@1=30\% and R@100=95\% is slightly lower than the prediction due to a couple of reasons. First, we find that Vivado did not achieve the target frequency of 140MHz, based on which the performance is predicted. The final frequencies are 126 MHz and 139 MHz, respectively. Second, PEs are connected by FIFOs, and full FIFOs at runtime can stall the pipeline. 
% Third, the number of vectors visited by the query set can be different from our mathematical estimation.
% Our model is still accurate considering the real/predicted performance is virtually the same after frequency normalization.
% The slight difference is either because (a) full FIFOs stall the pipeline or (b) the number of vectors visited at query time is different from the estimation.

\textbf{Third, FPGA performance is closely related to $K$, as instantiating longer priority queues consumes a lot of resources.} 
% As shown in Table~\ref{tab:final_designs}, 
To match the performance of Stage PQDist that contains many compute PEs, \textit{FANNS} needs to instantiate many hardware priority queues in Stage SelK. 
But the resource consumption per queue is roughly linear to the queue size $K$. As $K$ grows, more resource consumption on queues results in fewer resources for other stages and leads to overall lower performance. This explains why the FPGA performance is slightly surpassed by the CPU when $K=100$.

\textbf{Fourth, picking appropriate algorithm parameters is essential for performance, regardless of hardware platforms.}
The performance numbers of the CPU and the baseline FPGA designs show that the QPS difference can be as significant as one order of magnitude with different parameters. 
% Plus, different hardware favors different settings. 

% the throughput (QPS) and latency between \textit{FANNS} and the CPU/FPGA baselines in Figure~\ref{fig:fpga_cpu_throughput} and Figure~\ref{fig:fpga_cpu_latency}. 
% The throughput experiments allow query batching (size = 10K) to report the highest QPS, while the latency experiments disable batching to minimize the latency. 
% In terms of throughput, \textit{FANNS} reports $1.3\sim23.0\times$ QPS as the baseline FPGA designs and $0.8\sim37.2\times$ as the CPU. 
% In terms of latency, \textit{FANNS} achieves $2.4\sim5.2\times$ better latency than the best CPU baseline and $1.3\sim1.8\times$ speedup against the best FPGA baseline numbers. 
% Several observations from the experiments include:

\subsubsection{Online Query Processing and Scalability}
\hfill \\

To support low-latency online query processing, we integrate \textit{FANNS} with a hardware TCP/IP stack~\cite{100gbps}, such that clients can query the FPGA directly, bypassing the host server. We also compare system scalability of GPUs and FPGAs in this scenario. As the network stack also consumes hardware resources, we rerun the \textit{FANNS} performance model to generate the best accelerators. We assume the queries already arrive at the host server for CPU and GPU baselines, while for FPGAs, the measurements include the network latency (around five $\mu$s RTT). 

\textbf{FPGA achieves 2.0$\sim$4.6$\times$ better P95 latency than the best CPU baseline.} Figure~\ref{fig:fpga_cpu_latency} captures the latency distributions~\cite{hoefler2015scientific} of each type of hardware. Although showing high tail latency, GPUs still achieve lower median and P95 latency than FPGAs and CPUs due to the much higher flop/s and bandwidth.
The FPGA shows much lower latency variance than its counterparts, thanks to the fixed accelerator logic in FPGAs.

\textbf{FPGAs achieves 5.5$\times$ and 7.6$\times$ speedup over GPUs in median and P95 latency in an eight-accelerator setup, as shown in Figure}~\ref{fig:first_page_overview}.
We run the prototype scale-out experiments on a cluster of eight FPGAs. 
Each FPGA or GPU holds a 100-million vector partition, running the same index (\textit{nlist}=8192, $m$=16) to achieve R@10=80\%.
For FPGAs, we use a CPU server that sends queries to all FPGAs and aggregates the results. 
For GPUs, Faiss natively supports multi-GPU workload partitioning. 
FPGAs achieve better scalability thanks to their stable latency distribution, as shown in Figure~\ref{fig:fpga_cpu_latency}. In contrast, GPUs experience long tail latencies, thus a multi-GPU query is more likely to be constrained by a slow run.

\begin{figure}[t]
% 	\vspace*{-1em} % to shrink gap between figures
	% \vspace*{-5mm} % to shrink gap between figures
	% full width, can be adjusted
	\centering
  \includegraphics[width=1.0\linewidth]{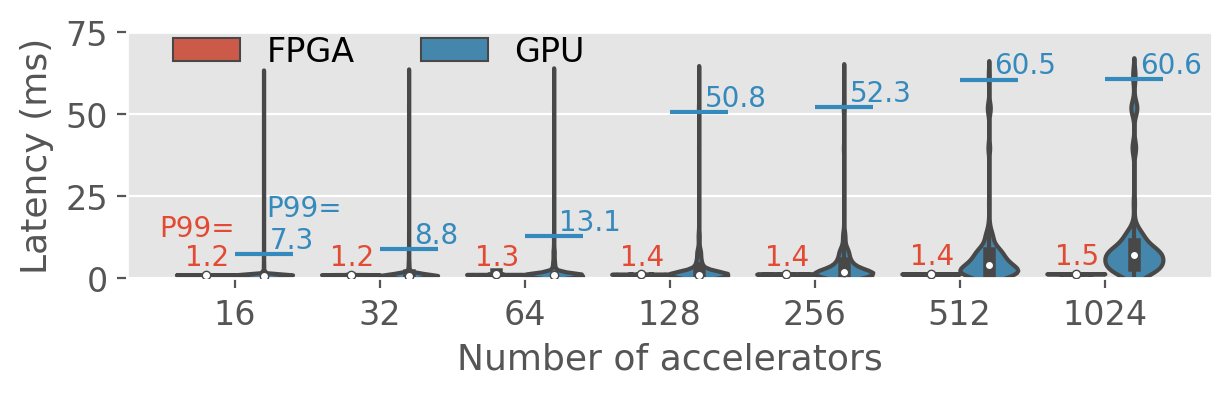}
	% \vspace{-2.5em} % to shrink gap between figures
 \vspace{-3em} 
  \caption{{Estimated latency on large-scale deployments.}}
  \label{fig:latency_simulation}
	\vspace{-1em} % to shrink gap between figures
\end{figure}

\textbf{FPGAs are expected to exhibit increasing speedup over GPUs as the search involves more (hundreds or thousands of) accelerators}.
To extrapolate latency trends beyond eight accelerators, we estimate the latency distribution of large-scale vector search using the following method. The query latency consists of search and network components. 
We record search latencies of 100K queries on a single FPGA/GPU using the same parameters as the above paragraph. For a distributed query, we randomly sample \textit{N\textsubscript{accelerator}} latency numbers from the latency history and use the highest number as the search latency. 
We assume the implementation of broadcast/reduce communication collectives follows a binary tree topology. Subsequently, we apply LogGP~\cite{culler1993logp, alexandrov1995loggp} to model the network latency, using previously reported values measured for InfiniBand using MPI~\cite{hoefler2014energy, hoefler2007low}: the maximum communication latency between two endpoints is 6.0 $\mu$s; the constant CPU overhead for sending or receiving a single message is 4.7 $\mu$s; and the cost per injected byte at the network interface is 0.73 ns/byte. We assume merging partial results from two nodes takes 1.0 $\mu$s.
As shown in Figure~\ref{fig:latency_simulation}, FPGA's P99 latency speedup over GPUs increases from 6.1$\times$ with 16 accelerators to 42.1$\times$ with 1024 accelerators, thanks to the low search latency variance on FPGAs. 

% Median latency speedup: [0.72730532 0.76773282 0.81626839 0.87858413 1.60054916 3.02456523
%  5.04189611]
% P95 latency speedup: [ 1.01782141  2.54856735  4.72096031  6.56805997  8.85863238 27.45998442
%  34.99215646]
% P99 latency speedup: [ 6.13134109  7.13766926 10.1844883  37.21463103 37.27821205 42.06890967
%  41.02700053]

% Later compare FPGA/CPU/GPU (in a single figure, only the best performance of 3 recalls). Say GPU is still the better one, but as expected. If one turn our FPGA to ASIC, we can surpass GPUs.

% R@1=0.3
% Speedup over CPU best baseline: 1.28 x
% Speedup over FPGA best baseline: 1.33 x
% Speedup over CPU worst baseline: 22.34 x
% Speedup over FPGA worst baseline: 20.79 x
% CPU: $1.3\sim22.3$
% FPGA: $1.3\sim20.8$

% \subsection{Implications for ASIC Designs}

% Although \textit{FANNS} leverages the reconfigurability of FPGAs to explore the design space of vector search accelerators, the strategy cannot be applied to ASICs as they cannot be reconfigured per use case. 
% We argue that ASIC designs for vector search should set alternative goals than FPGAs to avoid the shifting bottleneck problem. 

% \textbf{ASIC designs should only target large-scale vector search, support only a subset of the algorithm operators, and cooperate with a general-purpose processor.}
% Instead of targeting all size scales, ASIC designs should target large-scale vector search. In such scenario, the bottlenecks will shift to Stage PQDist and Stage SelK as many database vectors must be scanned. 
% Thus, the ASIC designs may only support the operators of those stages to provide fast and energy-efficient scan functionality. 
% As the index traversal are more lightweight compared to vector scan in the large-scale case, a CPU can handle the former workloads and send the scan requests to the ASIC. 
% % Seco
% % Thus, ASIC designs need to rethink the target use case. 

%% file: related_work.tex
\section{Related Work}

% \subsection{ANNS on Modern Hardware}

To our knowledge, \textit{FANNS} is the first hardware-algorithm co-design framework for vector search. 
We introduce related works below.

\textbf{Vector search on modern hardware.}
The most popular GPU-accelerated ANN library so far is Faiss developed by Meta~\cite{johnson2019billion}. 
The academia has also built several GPU-based ANNS systems~\cite{wieschollek2016efficient, chen2019robustiq, chen2019vector}. 
Google researchers accelerate exact nearest neighbor search on TPUs and show great performance on one-million-vector datasets~\cite{chern2022tpu}.
\citet{lee2022anna} propose a fixed ASIC design for PQ supporting arbitrary algorithm parameters.%, and the simulation shows significant speedup over GPUs. 
\citet{zhang2018efficient} implements a variation of the PQ algorithm on FPGAs and focuses on compressing large IVF indexes to fit it to BRAM.
% Yet it only reports 70\% of recall on the SIFT dataset while we report 95\% recall. Since their artifact and selected parameters are not publically available, it is hard to directly compare the performance with \textit{FANNS}. 
% Apart from accelerator-based solutions, researchers have also explored modern storage for large-scale ANNS. 
\citet{ren2020hm} stores full-precision vectors in non-volatile memory to scale up graph-based ANNS, while on-disk ANNS has to be careful with I/O cost~\cite{chen2021spann, jayaram2019diskann, lejsek2008nv}.

\textbf{Vector search in hardware accelerated machine learning systems.}
Vector search is an essential component in retrieval-augmented language models and recommender systems. For recommender systems, previous work has already explored hardware acceleration using a single FPGA~\cite{jiang2020microrec}, a cluster of FPGAs~\cite{zhu2021distributed}, and hybrid GPU-FPGA clusters~\cite{jiang2021fleetrec} in which the FPGA serves as a disaggregated memory node controlling SRAM, HBM, and DDR memory. FANNS can be integrated with them in a single system, enabling the entire candidate generation and click-through rate prediction recommendation pipeline being offloaded to specialized hardware thus allowing high throughput while achieving low service latency.

\textbf{Vector search algorithms.}
There are several ways to index the vectors to reduce the number of vectors visited per query. 
The first category, as introduced in Section~\ref{sec:background}, is IVF indexes~\cite{IVF, babenko2014inverted}.
% The first category is clustering-based methods (known as inverted-file indexes)~\cite{IVF, babenko2014inverted}. Each vector is assigned to the nearest cluster, and only a subset of clusters close to the query vectors is visited during a search. 
The second category is graphs~\cite{malkov2014approximate, malkov2018efficient, fu2017fast, sun2014srs}. The graph should connect nodes (vectors) by edges that present neighborhood relationships. The search starts from an entry point and iteratively traverses the graph until one of the termination conditions is met. 
The third category is Locality-Sensitive Hashing (LSH)~\cite{gionis1999similarity, datar2004locality, huang2021point, zheng2016lazylsh}.
Each vector is hashed into a bucket that, to some extent, reflects the spatial location of the vector. 
A search hashes the query and scans the respective bucket.
To increase the chance of finding the nearest neighbor, one can create several hash tables and ensemble the search results.
LSH is a popular topic for theorists as it provides probabilistic theoretical guarantees.
The fourth category is trees~\cite{beis1997shape, silpa2008optimised}. Though studied in the early days, tree-based structures cannot handle high-dimensional features well due to the curse of dimensionality. 
Apart from indexing methods, researchers also propose to quantize the vectors to reduce memory footprint and save bandwidth. 
Product quantization~\cite{PQ, OPQ} is the most popular quantization algorithm for ANN search.

%% file: conclusion.tex
\section{Conclusion}

Commercial search engines are driven by a large-scale vector search system operating on a massive cluster of servers.
We introduce \textit{FANNS}, a scalable FPGA vector search framework that co-designs hardware and algorithm.
Our eight-FPGA prototype demonstrates 7.6$\times$ improvement in P95 latency compared to eight GPUs, with our performance model indicating that this advantage will only increase as more accelerators are employed. 
The remarkable performance of \textit{FANNS} lays a robust groundwork for future FPGA integration in data centers, with potential applications spanning large-scale search engines, LLM training, and scientific research in fields such as biomedicine and chemistry.

% The paper takes advantage of the increasingly available FPGAs to accelerate ANNS. 
% Designing specialized accelerators for ANNS is challenging, because the many design choices on both the algorithm and hardware sides can significantly influence the FPGA performance. 
% To this end, we present \textit{FANNS}, an end-to-end ANNS framework that, given a user-provided recall requirement, automatically figures out the best design choices by performance modeling and generates the optimal FPGA accelerator.
% The experiments show the effectiveness of customizing the FPGAs per recall, as the fully customized accelerators achieve up to 20.79$\times$ QPS over a fixed FPGA design and up to 29.98$\times$ QPS over a 16-core CPU. 
% Given that the FPGAs are becoming pervasive in data centers and that the devices 
% Apart from the proposed solution, we expect that the wave of FPGA architecture innovation will unlock more potential of FPGA-accelerated ANNS: first, the increasing amount of device memory on FPGAs (half-terabytes on Enzian) enables cost-efficient ANNS scale up; second, the trend of fusing FPGAs with CPUs on the same chip will make FPGA-accelerated ANNS more seamless and more viable. 

%% file: appendix.tex
\newpage
\appendix
\onecolumn{

% All manuscripts submitted to the SC Technical Papers program must contain an AD Appendix. The AD Appendix describes the significant research products and other evidence needed for greater scientific rigor and transparency of the scientific conclusions of the manuscript. If a manuscript’s scientific findings have no significant supporting research products or require no additional evidence in support of its scientific conclusions, the AD Appendix can easily be generated to reflect this fact. That is, the AD form is mandatory but the author makes the determination whether additional evidence is needed for their particular scientific contribution.

% The AE Appendix is optional but authors are strongly encouraged to provide AE resources and workflows that reviewers can easily evaluate for availability, functionality, and their support for claimed key results. Even though the AD/AE review process strongly encourages publicly available artifacts, it is open to artifacts that are not.

% Role of AD/AE Appendix during review: Submissions to SC are double-blind reviewed. SC Papers Committee (PC) reviewers will have only the information in the AD/AE form available to them that does not compromise the double-blind protection.

% The AD/AE review criteria are posted as part of AD/AE Appendix Process & Badges. The AE badges are the rubric as to how far an artifact has gone along in the reproducibility spectrum (see description of the badges).

\section{Artifact Description}

We introduce \textit{FANNS}, an end-to-end and scalable vector search framework on FPGAs. Given a user-provided recall requirement on a dataset and a hardware resource budget, \textit{FANNS} automatically co-designs hardware and algorithm, subsequently generating the corresponding accelerator. The framework also supports scale-out by incorporating a hardware TCP/IP stack in the accelerator. 

We use the following open-source software and commercial hardware in the experiments.
The experimental results should be easily reproducible on the same hardware.

\subsection{Hardware.}

For CPU experiments, we use an m5.4xlarge CPU server on AWS. The server contains 16 vCPUs of Intel(R) Xeon(R) Platinum 8259CL @ 2.50GHz (Cascade Lake, 14nm technology) and 64 GB of DDR4 memory.
For GPU experiments, we use a p3.24xlarge GPU server on AWS. It contains eight NVIDIA V100 GPUs manufactured by the TSMC 12 nm FFN (FinFET NVIDIA) technology. Each GPU has 5,120 CUDA cores and 32 GB HBM2.
For FPGA experiments, we use Xilinx Alveo U55c FPGA fabricated with TSMC's 16nm process. Each FPGA contains 1.3M LUTs, 9K DSPs, 40MB on-chip memory, and 16 GB HBM. We use a cluster of FPGAs for scale-out experiments, and all the U55c FPGAs are connected to the same switch.

\subsection{Software.}

We use Faiss 1.7.1 for both CPU and GPU experiments. The CUDA version of the GPU server is 11.3.
For FPGA experiments, we develop the accelerators using Vitis HLS 2022.1.

\section{Artifact Evaluation}

% It includes:
% (i) a complete description of the experiment workflow that the computational artifact(s) can execute
% (ii) an estimation of the execution time to execute the experiment workflow.
% (iii) a complete description of the expected results and an evaluation of them, and most importantly.
% (iv) how the expected results from the experiment workflow relate to the results found in the article. Best practices indicate that, to facilitate the understanding of the scope of the reproducibility, the expected results from the artifact should be in the same format as the ones in the article. For instance, when the results in the article are depicted in a graph figure, ideally, the execution of the code should provide a (similar) figure (there are open-source tools that can be used for that purpose such as gnuplot). It is critical that authors devote their efforts on these aspects of the reproducibility of experiments to minimize the time needed for their understanding and verification.

The code of FPGA accelerators with/without the network stack as well as the CPU/GPU baseline evaluations are available at: \url{https://github.com/WenqiJiang/SC-ANN-FPGA}. We only apply for the availability badge because the HACC FPGA cluster at ETH Zurich does not allow anonymous access for reproducibility evaluation. 

The repository contains three folders with documented execution flow:

\begin{itemize}

    \item The \textit{CPU\_GPU\_baselines} folder contains all the CPU and GPU baseline experiments. It also contains part of the co-design formula -- we use Faiss to evaluate the relationship between index, \textit{nprobe}, and recall on a given dataset.

    \item The \textit{FPGA\_local} folder contains all the \textit{CD-ANN} programs without the network stack. Its \textit{performance\_model} sub-directory contains the performance and resource consumption models. A user can use the programs in it to predict the best hardware design given a certain recall goal on a dataset. The \textit{code\_generation} sub-directory can generate the ready-to-compile FPGA code given a hardware setting, which is either predicted by the performance model or set by the user manually. The \textit{generated\_projects} sub-directory contains all the generated FPGA accelerator code that we evaluated in the experiments. 
    
    \item The \textit{FPGA\_with\_network} folder contains all the FPGA accelerators that we evaluated with the network stack. The FPGA kernels are located in the \textit{kernel/user\_kernel} sub-directory. The CPU client programs are in the \textit{CPU\_program} sub-directory.
\end{itemize}

It will take a significant amount of time to reproduce all results. Training the indexes can be costly (around two hours per index), given that we trained 18 indexes in the paper. Once the indexes are trained, it should take less than two hours on CPUs and GPUs to reproduce the performance. The FPGA compilation can take around ten hours per design, which is costly because we have more than ten designs to form the experiments. Executing an FPGA program should take less than a minute.

The reproduced results are expected to be identical to the paper, given that we already reported performance on multiple runs and characterized the error bars in throughput experiments as well as latency distributions in latency experiments.

}